\documentclass[sigconf, nonacm]{acmart}

\AtBeginDocument{%
	}

\usepackage{graphicx}
\usepackage{amsmath}
\usepackage{multirow}
\usepackage{booktabs}
\usepackage{color, soul, xcolor}
\usepackage{balance}

\begin{document}
\begin{sloppypar}

\title{Blind Image Super-resolution with Rich Texture-Aware Codebooks}
\author{Rui Qin$^{\dag}$, Ming Sun$^{\ddag}$, Fangyuan Zhang$^{\dag}$, Xing Wen$^{\ddag}$, Bin Wang$^{*,\dag}$\\
$^{\dag}$School of Software, Tsinghua University \& BNRist, China, $^{\ddag}$ Kuaishou Technology, China}

% \author{Rui Qin}
% \affiliation{%
%   \city{Beijing}
% \additionalaffiliation{%
%   \institution{Beijing National Research Center for Information Science and Technology}
%   \city{Beijing}
% \email{qr20@mails.tsinghua.edu.cn}

% \author{Ming Sun}
% \affiliation{%
%   \institution{Kuaishou Technology}
%   \city{Beijing}
% \email{sunming03@kuaishou.com}

% \author{Fangyuan Zhang\footnotemark[1]}
% \affiliation{%
%   \city{Beijing}
% \email{zhangfy19@mails.tsinghua.edu.cn}

% \author{Xing Wen}
% \affiliation{%
%   \institution{Kuaishou Technology}
%   \city{Beijing}
% \email{td.wenxing@gmail.com}

% \author{Bing Wang\footnotemark[1]}
% \authornote{Cooresponding author. This work was supported by the National Natural Science Foundation of China under Grant 62072271.}
% \affiliation{%
%   \city{Beijing}
% \email{wangbins@tsinghua.edu.cn}

\thanks{
  $^{*}$ Corresponding authors.
}
\thanks{This work was supported by the National Natural Science Foundation of China under Grant 62072271.}
\thanks{
 Rui Qin: qr20@mails.tsinghua.edu.cn,\\
 Ming Sun: sunming03@kuaishou.com, \\
 Fangyuan Zhang: zhangfy19@mails.tsinghua.edu.cn, \\
 Xing Wen: td.wenxing@gmail.com, \\
 Bin Wang: wangbins@tsinghua.edu.cn}

% !TEX root = sample-xelatex.tex
\begin{abstract}
    Blind super-resolution (BSR) methods based on high-resolution (HR) reconstruction codebooks have achieved promising results in recent years. However, we find that a codebook based on HR reconstruction may not effectively capture the complex correlations between low-resolution (LR) and HR images. In detail, multiple HR images may produce similar LR versions due to complex blind degradations, causing the HR-dependent only codebooks having limited texture diversity when faced with confusing LR inputs. To alleviate this problem, we propose the Rich Texture-aware Codebook-based Network (RTCNet), which consists of the Degradation-robust Texture Prior Module (DTPM) and the Patch-aware Texture Prior Module (PTPM). DTPM effectively mines the cross-resolution correlation of textures between LR and HR images by exploiting the cross-resolution correspondence of textures. PTPM uses patch-wise semantic pre-training to correct the misperception of texture similarity in the high-level semantic regularization. By taking advantage of this,  RTCNet effectively gets rid of the misalignment of confusing textures between HR and LR in the BSR scenarios. Experiments show that RTCNet outperforms state-of-the-art methods on various benchmarks by up to \textbf{\textit{0.16 $\sim$ 0.46dB}}.
\end{abstract}


\renewcommand{\authors}{Rui Qin, Ming Sun, Fangyuan Zhang, Xing Wen, Bin Wang}
\renewcommand{\shortauthors}{Rui Qin, Ming Sun, Fangyuan Zhang, Xing Wen, Bin Wang}
\maketitle

% !TEX root = sample-xelatex.tex
\section{Introduction}
\label{sec:intro}

Blind Super-Resolution (BSR) 
	aims to realistically reconstruct high-resolution (HR) images from low-resolution (LR) images with unknown degradation~\cite{real-esrgan,DAN,CDC,femasr,swinir,QuanTexSR,li2020blind}.
To avoid the General Adversarial Network (GAN) artifact, 
	codebook-based BSR approaches~\cite{QuanTexSR,femasr}, inspired by VQVAE~\cite{vqvae,vqvae2} and VQGAN~\cite{vqgan}, model high-resolution textures using a discrete feature space created by a pre-trained feature codebook to reconstruct HR images.
These methods
	 have shown promising results, as the codebook effectively constrains the output to fall within a valid solution space.

\begin{figure}
	\includegraphics[width=\columnwidth]{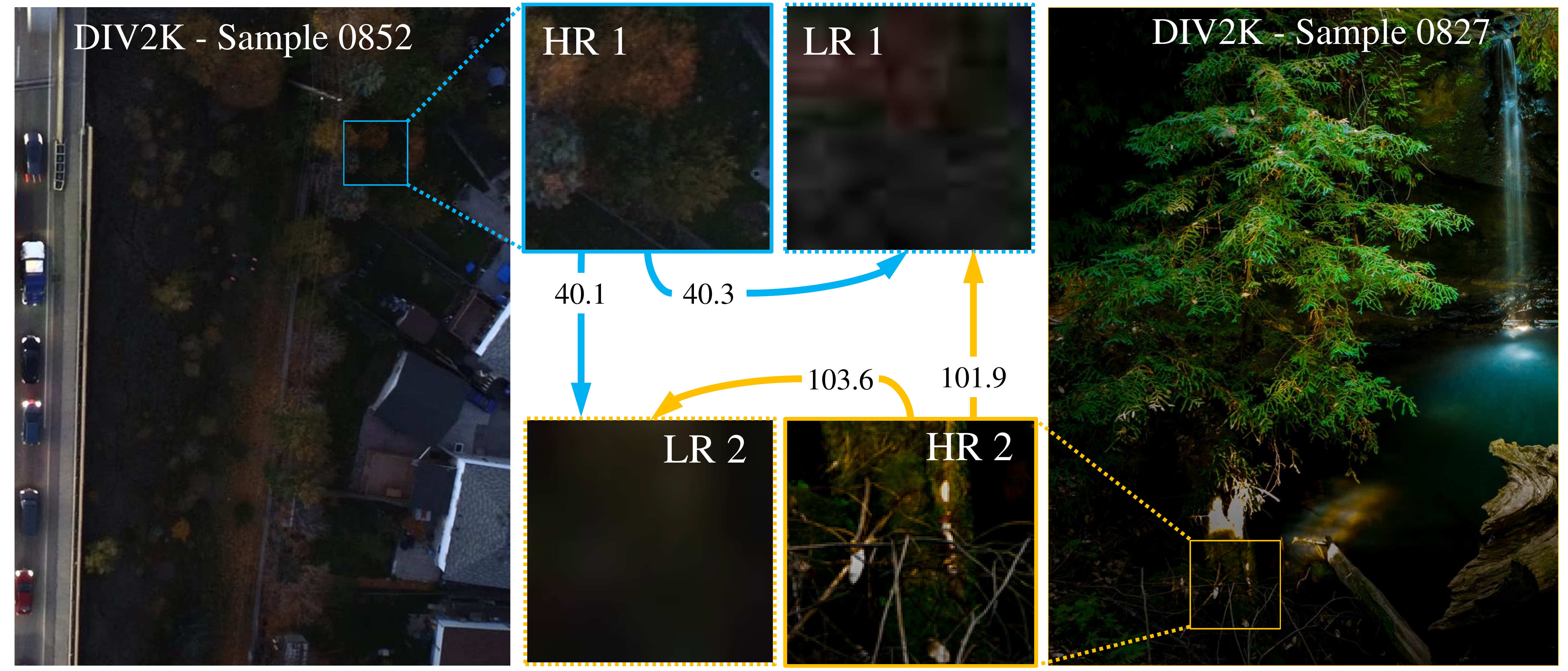}
	\caption{Confusing LR samples with different HR textures processed with the same random degradation used in \cite{BSRGAN,QuanTexSR,femasr} (including various noise, blur, and compression). The MSE in RGB space on the line indicates the patch distance.}
	\label{fig: confuse_lr}
\end{figure}

\begin{figure*}[htbp]
	% \vspace{-0.2in}
	\begin{center}
	\includegraphics[width=\linewidth]{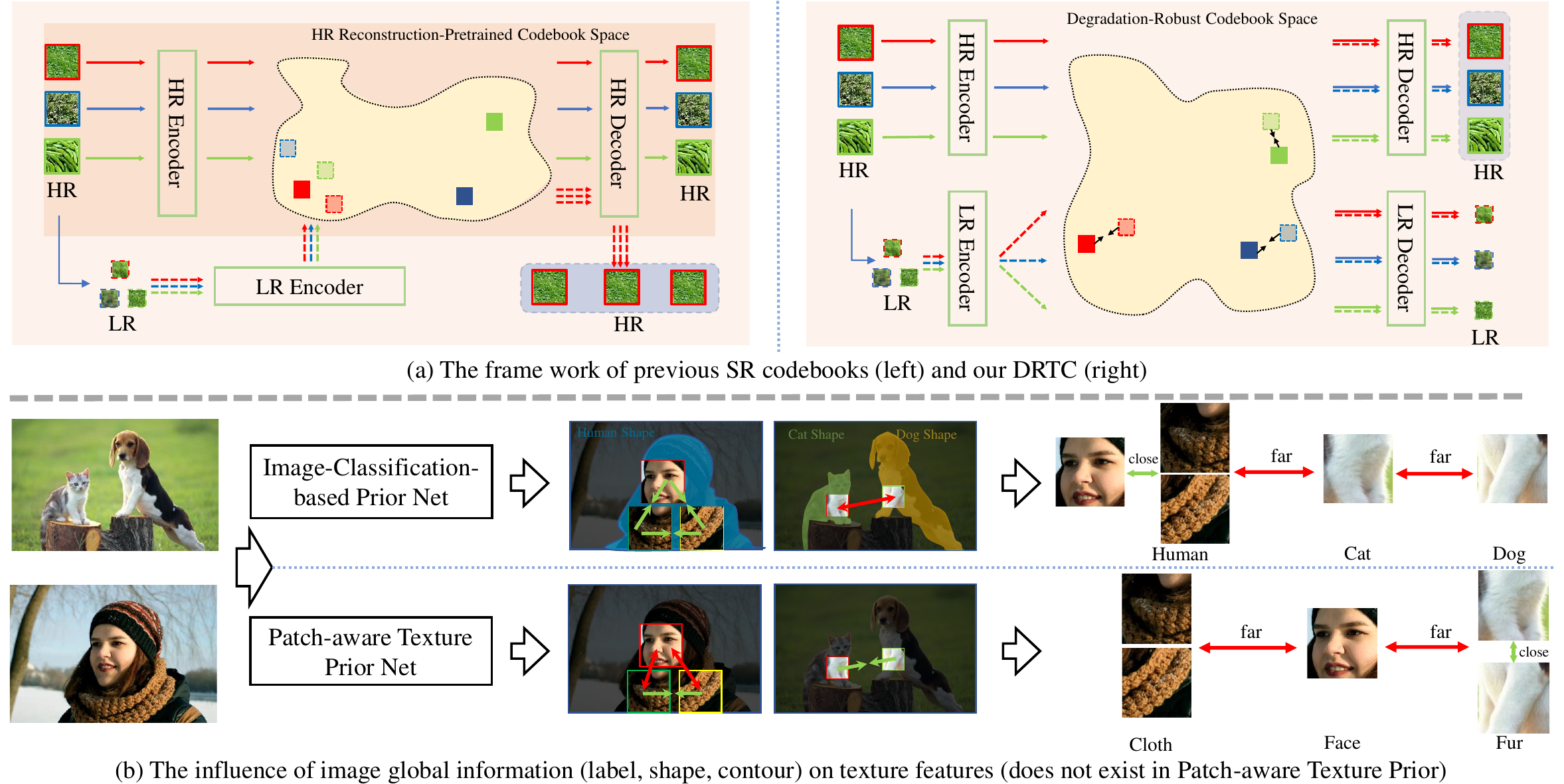}
	% \vspace{-0.1in}
	\caption{An illustration of our motivation. \textit{(a)} Left: The previous HR-reconstruction-based codebook trained only on HR reconstruction, requiring a second stage training for the LR encoder; right: DTPM, which incorporates both resolutions and cross-resolution consistency during training. \textit{(b)} Top: Image classification-based features susceptible to global factors such as class labels, object shapes, and contours; bottom: our PTPM prior without the influence of global information.}
	\label{fig: page2_idea}
	\end{center}
	% \vspace{-0.1in}
\end{figure*}

One of the major challenges in BSR
	 is the complex blind degradation, which leads to similar LR versions from different HR inputs, disrupting the LR-HR matching correlation~\cite{DAN,CDC,real-esrgan,swinir,femasr,QuanTexSR}.
For example, 
	in Fig.~\ref*{fig: confuse_lr}, we sample two HR images from the DIV2K~\cite{DIV2K} dataset and degrade them using the widely used blind degradation procedure of ~\cite{BSRGAN}. We compute the Mean Square Error (MSE) for the similarity evaluation and find that complex degradation reduces the distinction between LR patches compared to their HR distinction. 
In detail, 
	HR 1 has a smaller MSE (40.1) with LR 2, in contrast to its own corresponding LR patch (LR 1), which has an MSE of 40.3. In addition, similar LR patches tend to match the same HR patch rather than their individual HR versions. Such phenomena complicate the handling of LR data.

To address this issue, recent codebook-based methods~\cite{QuanTexSR,femasr} incorporate an additional LR encoder to model the LR-HR relationship, based on the texture codebook learned from HR data (Fig.~\ref{fig: page2_idea}.a).
While this technique is effective when dealing with mildly degraded data, it shows lower-quality results when handling severely degraded areas. We find two main factors that limit its performance on complex degraded data.
\textbf{\textit{1)}}
	First, the previous codebook space, built from distinct HR data, struggles with confusing LR inputs. Unlike the clear relationship between HR textures, HR-LR relationships within BSR are confusing and often many-to-one. This poses a challenge for the previous codebooks pre-trained for HR reconstruction~\cite{QuanTexSR,femasr} to distinguish different textures from similar degraded versions, thus limiting the diversity of texture restoration. Besides, to simplify learning, they apply the codebook only at the network bottleneck, which effectively captures larger textures but may miss mid-to-low-level details.
\textbf{\textit{2)}} 
	Second, they use image classification-based features (often from backbones like VGG~\cite{vgg} pre-trained on ImageNet) for additional semantic regularization during codebook learning. 
	However, high-level tasks that prioritize global semantics may neglect local information~\cite{jiang2021robust,li2022mage} crucial for low-level tasks, causing inconsistency between pre-trained features and local texture perception (\textit{e.g.}, Fig.~\ref{fig: page2_idea}.b). 
	To this end, developing a texture-friendly and efficient prior based on existing global prior is worthwhile for BSR tasks, but remains underexplored.

To address the first limitation, 
	we propose the Degradation-robust Texture Prior Module (DTPM). Unlike previous methods that rely solely on HR data for codebook learning (Fig.~\ref{fig: page2_idea}.a), DTPM involves LR data in codebook learning, improving the adaptability of codebooks to LR data. Furthermore, we exploit the distinguishability of HR representations to improve LR distinguishability by delving deeper into HR-LR correlation. Specifically, we enforce the consistency of paired HR-LR representations in codebook space and the consistency of texture content across resolutions in reconstruction results. Besides, we conduct a hierarchical codebook and a deep-to-shallow sequence training strategy for fine-grained texture modeling and stable optimization.
To address the second problem, 
	we propose the Patch-aware Texture Prior Module (PTPM) to improve the local texture perception of priors based on existing image labels. Specifically, we propose a patch-level classification-based pre-training task to reduce the global contour and shape influences. Simultaneously, we reorganize texture-friendly labels based on coarse feature clustering to correct the misleading feature similarity caused by global labels. Feature visualization and ablation studies show that PTPM offers better texture similarity assessment and benefits subsequent BSR tasks.
\begin{figure*}[t]
    \includegraphics[width=0.88\textwidth]{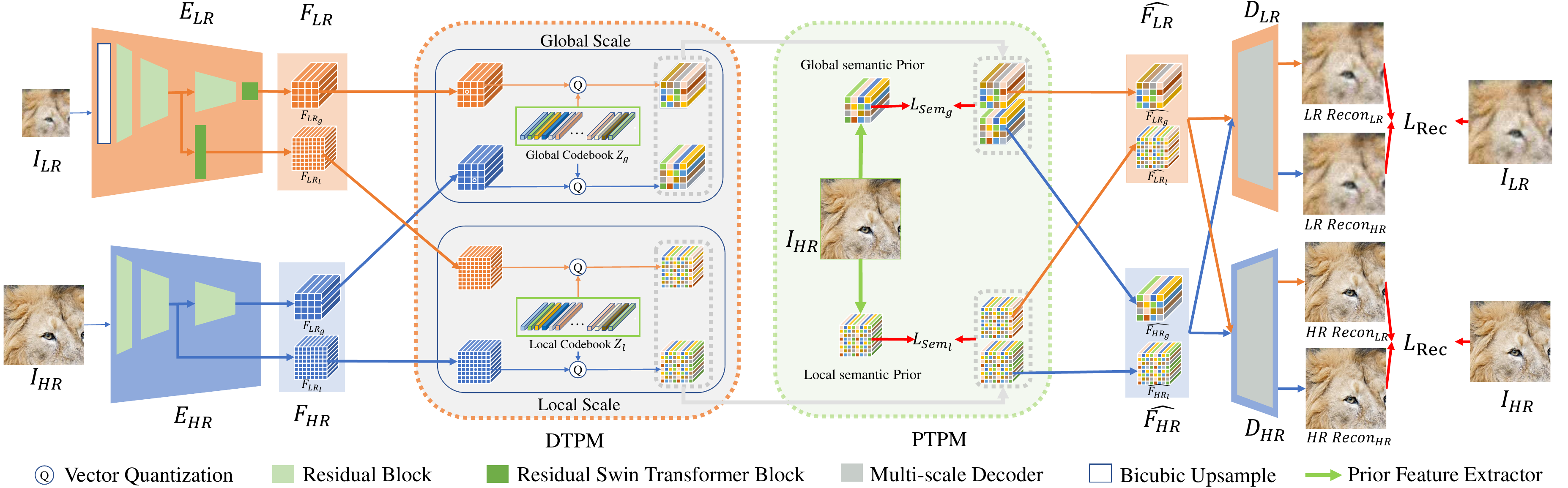}
    % \vspace{-0.1in}
    \caption{The RTCNet framework. 
	\textbf{(1)} 
	During training, LR and HR input images are encoded using multi-scale encoders. These features are quantized in multi-scale codebooks via DTPM. 
	The LR and HR decoders then perform dual-resolution reconstruction.
	\textbf{(2)} 
	During inference, only LR images are used as input; these are processed by the LR encoder and DTPM to obtain multi-scale quantized features, which are then used by the HR decoders to reconstruct super-resolution images.
	}
    \label{fig:pipeline}
	% \vspace{-0.1in}
\end{figure*}
By integrating DTPM with PTPM, we propose the Rich Texture-aware Codebook-based Network (RTCNet) for BSR. Experiments on several benchmark datasets show that RTCNet outperforms state-of-the-art methods by up to 0.16 $\sim$ 0.46dB in PSNR and provides competitive perceptual performance.
Our contributions are summarized as follows:
\begin{enumerate}
	\item To alleviate the limitations of previous codebook-based methods in modeling texture diversity and granularity, we propose the Degradation-robust Texture Prior Module (DTPM), which incorporates the cross-resolution consistency and hierarchical codebook structure of the texture.
	\item We propose the Patch-aware Texture Prior Module (PTPM). Compared to previous image classification-based priors, PTPM eliminates the influence of global information on local texture learning by patch-level pre-training with texture-friendly reorganized labels.
	\item Compared to recent methods, the proposed RTCNet framework combining DTPM and PTPM achieves state-of-the-art performance on multiple benchmark datasets. 
\end{enumerate}

% !TEX root = sample-xelatex.tex
\section{Related Work}
\label{sec:related}
\subsection{Codebook-based SISR}
Traditional codebook-based methods~\cite{yang2010image, chang2004super} have been effective in modeling low-resolution (LR) and high-resolution (HR) patches in color spaces, especially under light degradation.
However, in the case of blind super-resolution (BSR) with severe and unknown degradation, their effectiveness decreases due to the complex correspondence of different resolutions.
Recent advances in deep learning~\cite{vqgan,vqvae,vqvae2,niu2020single} have enabled the development of vector quantization-based methods~\cite{QuanTexSR,femasr,zhou2022towards} that transition patch matching from pixel to feature space, showing notable improvements in BSR scenarios. Specifically, these methods used a high-resolution VQVAE~\cite{vqvae,vqvae2} generation model (vector codebook and decoder) to model HR textures and an additional LR encoder for cross-resolution feature matching. Despite these advantages, as mentioned in Sec.~\ref{sec:intro}, recent codebook-based methods still struggle with limited diversity and limitations and coarse modeling for fine textures. Therefore, we design the DTPM to alleviate codebook collapse and achieve hierarchical texture modeling.
\subsection{Prior-based SISR}
Since SR is inherently an ill-posed problem, using additional image priors can effectively improve the restoration performance. The prior-based super-resolution methods can be simply divided into explicit and implicit methods. Explicit methods~\cite{zheng2018crossnet, zhang2019image, yang2020learning,jiang2021robust, zhou2020cross,li2020blind,li2020enhanced,li2018learning}, which use HR reference images, can restore realistic textures but have low performance with limited reference images. Implicit methods~\cite{menon2020pulse, karras2020analyzing, gu2020image, pan2021exploiting} use generative model-based priors~\cite{stylegan} and  achieve superior results on domain-specific images such as faces~\cite{chan2021glean, wang2021towards, yang2021gan}. Several methods learn a posterior distribution with pre-trained StyleGAN~\cite{stylegan} and use another encoder to project LR images into StyleGAN's latent space. However, since learning the prior from the generative model on generic images is challenging, recent methods use the high-level task-based priors for image texture reconstruction~\cite{wang2018sftgan, QuanTexSR, femasr}.
However, they tend to overlook local textures and instead focus on global semantics, making them less suitable for texture-sensitive image restoration.
% !TEX root = sample-xelatex.tex

\section{Method}
\label{sec:method}

\subsection{Overview}
The framework of our method is shown in Fig.~\ref{fig:pipeline} and briefly described herein.

% \vspace{-0.1in}
\paragraph{\textbf{Training}}
During training, RTCNet inputs both low-resolution (LR) images $I_{LR}$ and high-resolution (HR) images $I_{HR}$, with CNN encoders $E_{HR}, E_{LR}$ used to extract hierarchical features $F_{HR},F_{LR}=E_{HR}(I_{HR}), E_{LR}(I_{LR})$ respectively.
Following prior work~\cite{femasr}, additional RSTB~\cite{swinir} layers are added to $E_{LR}$ for stronger learning ability. The extracted features are quantized via hierarchical codebooks in the Degradation-robust Texture Prior Module (DTPM),
\begin{equation}
	\begin{aligned}
	\widehat{F_{HR}} = DTPM(Z,F_{HR}),\widehat{F_{LR}} =  DTPM(Z,F_{LR})
	\end{aligned}
\end{equation} 
where $Z$ denotes the hierarchical codebooks in DTPM.
Finally, we pair the quantized features of all resolutions with decoders $D_{HR},D_{LR}$ for cross-resolution reconstruction, and compute the reconstruction loss against ground truth images.
\paragraph{\textbf{Inference}}
Different from training, we only apply the LR input to the HR reconstruction process to obtain the super-resolution result $I_{SR}$:
\begin{equation}
	\begin{aligned}
		I_{SR} = D_{HR}(DTPM(Z,E_{LR}(I_{LR})).
	\end{aligned}
\end{equation}
\vspace{-0.2in}
\subsection{Degradation-robust Texture Prior Module}
In this section, we introduce our Degradation-robust Texture Prior Module in detail, including vector quantization, cross-resolution consistency constraints, and the hierarchical codebook structure.
% \vspace{-0.1in}
\paragraph{\textbf{Vector Quantization}} 
For each point feature $f \in R^{C}$, its quantized result $\widehat{f} \in R^{C}$ is the nearest neighbor based on $L2$ distance in the codebook $Z\in R^ {N\times C}$ as
\begin{equation}
	\widehat{f} = Q(Z,f) = z_m, m= \arg\min_{j  \in [1,N]  }||f-z_j||_2,
\end{equation}
where N denotes the size of the codebook.
 Given the input feature $F\in R^{H\times W \times C}$, its quantized feature $\widehat{F}$ is the combination of the quantized results of all point features within $F$, expressed as $\widehat{F} = Q(Z,F)=\{Q(Z,f_{i,j})| i \in [1,H], j \in [1,W]\}$.
Following preious work~\cite{van2017neural, esser2021taming}, we directly copy the gradient from $\widehat{F}$ to $F$ for backpropagation and use the following loss function, $L_{Code}$ to optimize the codebooks:
 \begin{equation}
 	L_{Code}(F, \widehat{F}) = ||\widehat{F}-sg(F)||_2 + \beta \cdot ||sg(\widehat{F})-F||_2,\\
 \end{equation}
where $sg(\cdot)$ means stop-gradient operation and $\beta=0.25$~\cite{van2017neural, esser2021taming}.

In training, DTPM quantizes HR and LR features simultaneously. 
Its loss, $L_{DTPM}$, is the sum of $L_{code}$ of hierarchical codebooks:
 \begin{equation}
 	\begin{aligned}
	 	\!\!\!\!\!L_{DTPM} &= L_{Code}(F_{HR}, \widehat{F_{HR}})+L_{Code}(F_{LR}, \widehat{F_{LR}}).
 	\end{aligned}
 \end{equation}
 \vspace{-0.2in}
\paragraph{\textbf{Cross-Resolution Correlation Constraints}}
We investigate the texture correlation between HR and LR images, focusing on the cross-resolution consistency. We decompose the texture consistency between LR and HR data into two separate components: \textbf{\textit{1)} Reconstruction consistency constraint in RGB space.}  The similar code representations should have similar texture content across resolutions. 
Since paired HR and LR images share the same content, their quantized features $\widehat{F}_{HR}$ and $\widehat{F}_{LR}$ should be able to reconstruct both $I_{LR}$ and $I_{HR}$ inputs using decoders of both resolutions,
\begin{equation}
	\begin{aligned}
   \!\!\!\!\!\!\!\!LR_{recon_{LR}} &= D_{LR}(\widehat{F_{LR}}), LR_{recon_{HR}} = D_{LR}(\widehat{F_{HR}}), \\
   \!\!\!\!\!\!\!\!HR_{recon_{LR}} &= D_{HR}(\widehat{F_{LR}}), HR_{recon_{HR}} = D_{HR}(\widehat{F_{HR}}).
   \end{aligned}
\end{equation}
Generated images should align with their corresponding resolution inputs, to which image reconstruction supervision is applied,
\begin{equation}
	\begin{aligned}
		L_{Rec\ Con} = \sum_{i=\{LR,HR\}} L_{Rec}(I_{LR},LR_{Recon_{i}}) +L_{Rec}(I_{HR},HR_{Recon_{i}}),
	\end{aligned}
\end{equation}
where $L_{Rec}$ denotes the image reconstruction loss function in Sec~\ref{sec: loss}. 
\textbf{\textit{2)} Representation consistency constraint in codebook space.} Images with similar texture content across resolutions should have similar representations in the codebook space. Specifically, we constrain the features extracted from paired HR-LR images to be consistent with each other,
\begin{equation}
	L_{Rep\ Con} = ||F_{HR} - F_{LR}||_2.
\end{equation}

\begin{figure}[t]
	% \vspace{-0.1in}
	% \includegraphics[width=0.98\columnwidth]{./multi-scale.pdf}
	\includegraphics[width=1\columnwidth]{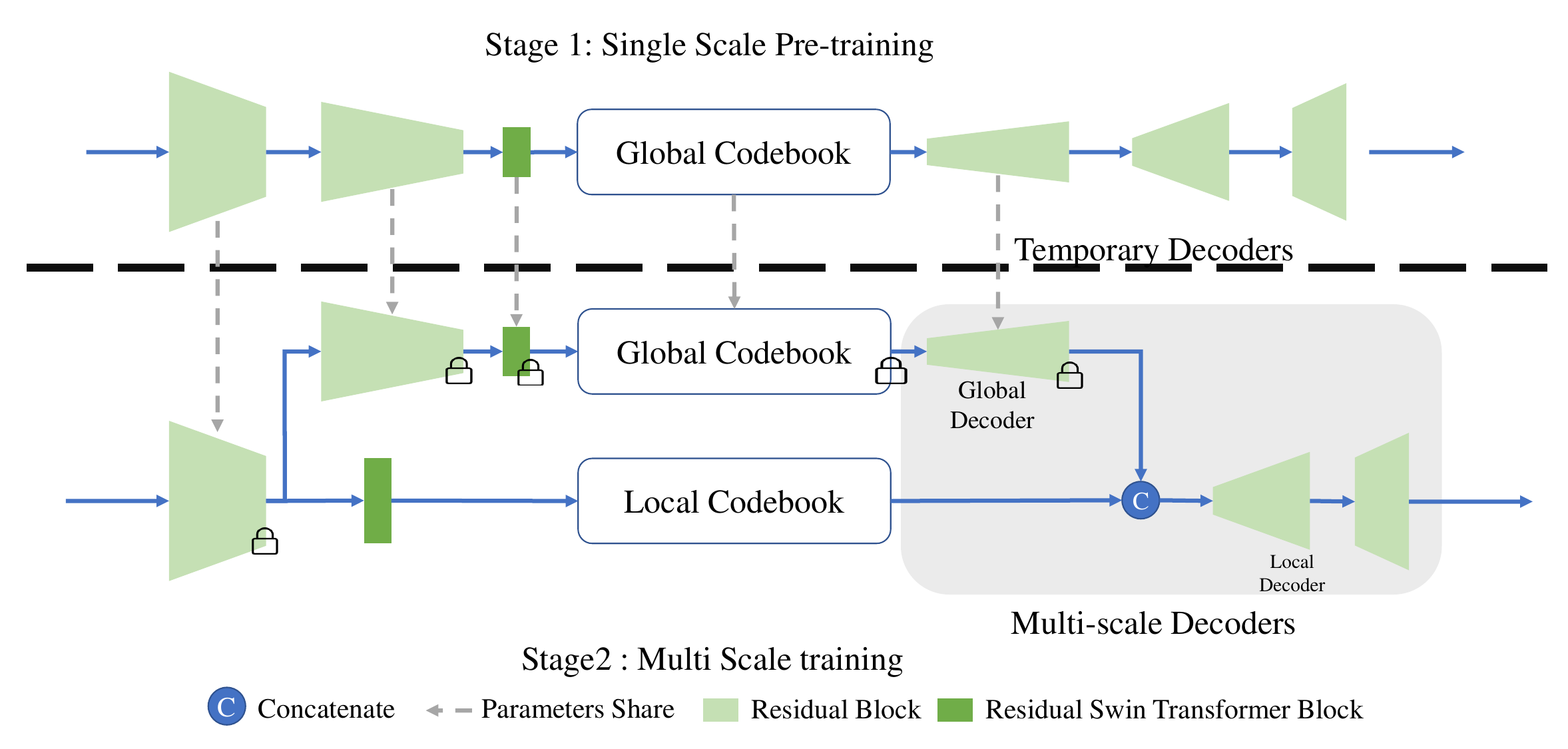}
	\caption{
		Hierarchical structure and its training strategy, using LR parts as an example due to the symmetry between LR and HR pipelines except for the RTSB in LR Encoder.}
	\label{fig:multi-scale}
%  \vspace{-0.2in}
\end{figure}

\paragraph{\textbf{Multi-scale Codebook Structure}}
The hierarchical codebook structure is based on the assumption that textures of different sizes can be characterized by codebooks of different scales. In the implementation, 
we employ two scales of $\times4$ and $\times8$ downsampling, hereafter referred to as local scale $l$ and global scale $g$ below. We apply codebooks to these scales for feature quantization,
\begin{equation}
	\begin{aligned}
	\!\!\!\!DTPM(Z,F_{HR}) = \{Q(Z_{g},F_{HR_{g}}), Q(Z_{l}, F_{HR_{l}}) \}, \\
	\!\!\!\!DTPM(Z,F_{LR}) = \{Q(Z_{g},F_{LR_{g}}), Q(Z_{l}, F_{LR_{l}}) \}. \\
	\end{aligned}
\end{equation} 
In contrast to previous bottleneck-like methods, additional shallow codebooks can represent diverse and minute texture information at smaller scales, which is helpful for generating finer textures.
To mitigate convergence difficulties when training multiscale codebooks from scratch, we propose a deep-to-shallow training strategy. Specifically, codebooks are trained sequentially, starting from the deepest scales and progressing toward the shallowest scales.
Fig.~\ref{fig:multi-scale} shows the detailed training strategy. First, the global codebook is trained starting from scratch, and the temporary decoder is implemented in place of the local codebook and the multi-scale decoders. In this phase, the multi-scale encoder and the global codebook are trained well. Second, we introduce the local codebook and replace the temporary decoder with multi-scale decoders, and freeze the well-trained modules in Stage 1.
The well-trained encoder and the global codebook allow for more effective and stable optimization of the local codebook.

\begin{figure*}[t]
	\includegraphics[width=\linewidth]{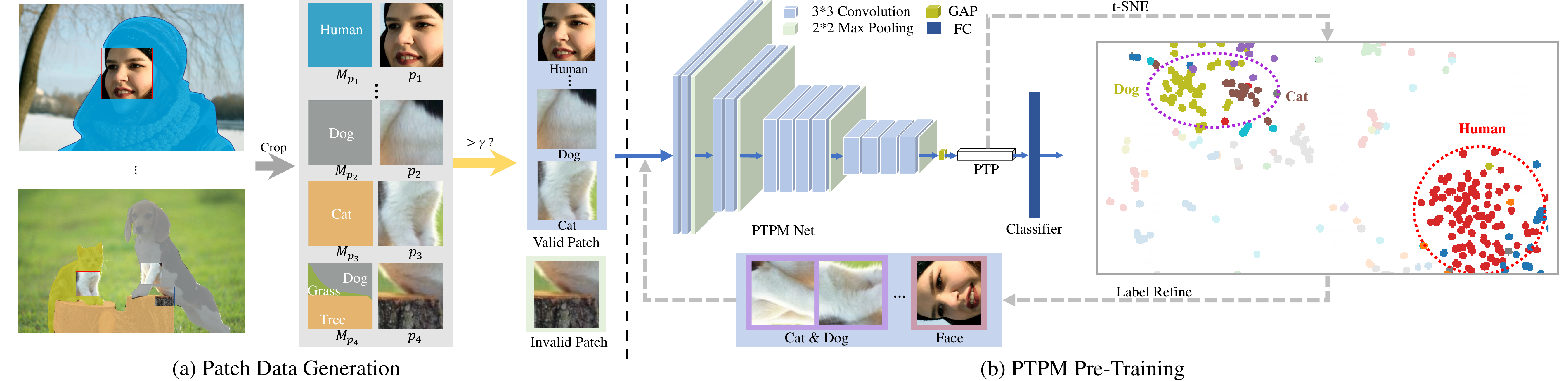}
    % \vspace{-0.1in}
	\caption{PTPM consists of two main blocks: (a) patch data generation; (b) patch classification training and label refinement.}
	\label{fig:low-level-prior idea}
    % \vspace{-0.2in}
\end{figure*}

\subsection{Patch-aware Texture Prior Module}
\label{Sec:low-level prior}
To obtain a low-level friendly prior emphasizing local details over global semantics for enhanced texture perception (Sec.~\ref{sec:intro}), we build our Patch-aware Texture Prior Module (PTPM) upon patch-level classification pre-training, drawing insights from multiple instance learning~\cite{pathak2014fully, poria2015deep} and fine-grained image classification~\cite{hou2016patch}. This section details the creation process of PTPM, covering data generation and agent task pre-training.

% \vspace{-0.1in}
\paragraph{\textbf{Patch Data Generation}}
In general, non-overlapping patches are extracted from the images, and those without sufficient segmentation labels are discarded.  The rest are assigned respective labels. Fig.~\ref{fig:low-level-prior idea}(a) illustrates the process of extracting a patch $p\in R^{H_p \times W_p \times 3}$ from an image $I$ with segmentation map $M$ in a non-overlapping manner. For each patch $p$, we consider it valid and assign its category label $Y_p$ as $y \in Y$, if the proportion of $M_p$ belonging to $y$ exceeds $\gamma$. Otherwise, it is deemed invalid,
\begin{equation}
	\label{eq6}
	VALID(p,M_p, Y)=\left\{
	\begin{aligned}
	1 &, & \exists \ y \in Y,\frac{\sum (M_p == y)}{H_{p} \times W_{p}} > \gamma, \\
	0 &, & \forall \ y \in Y,\frac{\sum (M_p == y)}{H_{p} \times W_{p}} \leq \gamma. \\
	\end{aligned}
	\right.
	% \vspace{-0.1in}
\end{equation}
The patch-category data pairs $P=\{(p,c)|VALID(p,c) = 1 \}$ are collected as the dataset for training the PTPM net. In the implementation, we manually set the threshold $\gamma = 0.85$ to balance data quality and quantity. This results in 16,818 effective patches across 27 classes, with 15,056 training and 1,763 validation samples.
Dividing the images into patches allows for the separation of different texture classes. For instance, a cat or dog's head represents a characteristic patch within its class, while the body parts show inter-class similarities. As the data is cropped into patches, classifying inter-class similar patches becomes more challenging, allowing easier grouping of similar patches with different class labels while increasing the distance between class-unique and inter-class similar patches at the same time. Therefore, patch-level pre-training allows for improved texture aggregation at the patch level.

% \vspace{-0.1in}
\paragraph{\textbf{Patch Classification Pre-training}}
We perform patch-level classification on the collected data, as shown in Fig.~\ref{fig:low-level-prior idea}.b. Specifically, we use the CNN part of VGG19 before the 3rd max pooling layer as our PTPM net $\phi_{patch}$, pre-initialize with ImageNet pre-trained weights, and add an additional linear classifier $C$ for pre-training. To ensure compatibility of the learned prior with the $L2$ distance in the codebook space, we add contrastive supervision $L_{Info NCE}$~\cite{infonce} to the cross-entropy loss $L_{CE}$. Patches within the same category are treated as positive samples, while patches from different categories are negative samples in $L_{Info NCE}$. Given patch samples $P=\{p_i|i=0,1,...,k\}$ and labels $Y=\{y_i|i=1,2,3,...,k\}$, the total prior training loss function is:
\begin{equation}
	\begin{aligned}
		L_{prior} &= L_{CE} + \lambda L_{Info NCE}\\
		&=-\sum_{i =1}^{k} y_i \log \left(\hat{y_i}\right) - \lambda \log \frac{\exp \left(||q_i-q_{i_{+}}||_2 / \tau\right)}{\sum_{i=1}^k \exp \left(||q_i-q_{i}||_2 / \tau\right)},
	\end{aligned}
\end{equation}
where $q_i=\phi_{patch}(p_i)$ denotes the feature embedding of $p_i$ after GAP and $\hat{y_i}=C(q_i)$ denotes the prediction results.

\begin{figure}[t]
	\includegraphics[width=1\columnwidth]{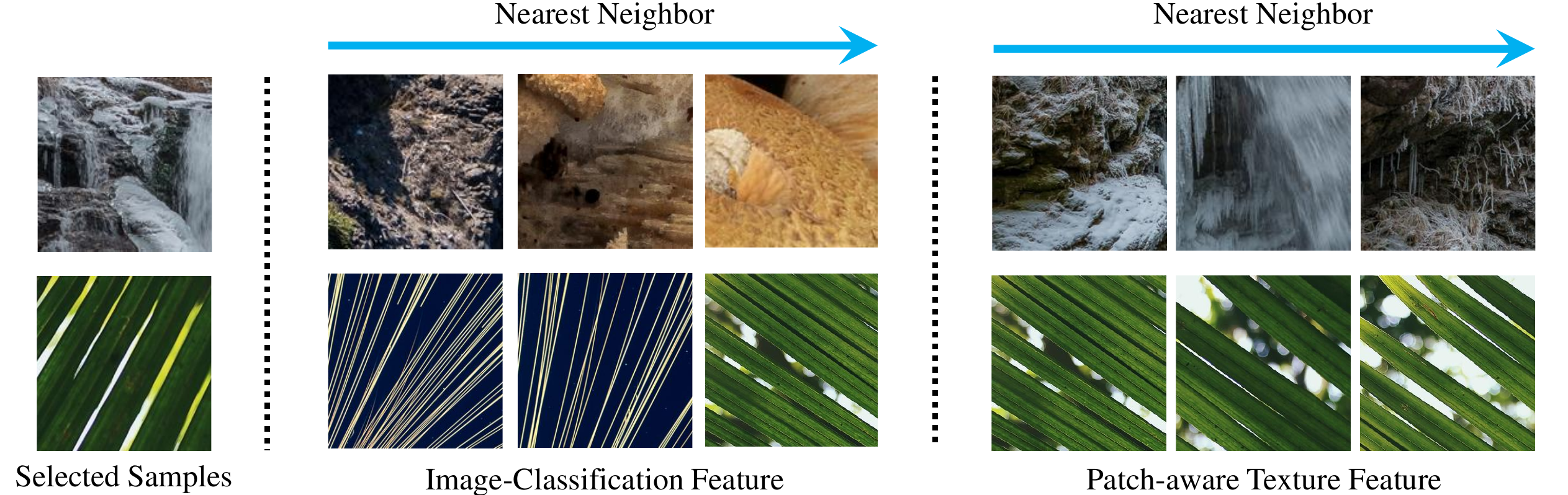}
	% \vspace{-0.1in}
	\caption{L2 nearest neighbors of several selected samples in different prior spaces.}
	\label{fig:low-level-prior visualize}
    % \vspace{-0.2in}
\end{figure}

% \vspace{-0.1in}
\paragraph{\textbf{Texture-orient Label Reorganization and Prior Refinement}}
Coarse pre-training using patch data and original image-level labels may be affected by global label influence. We mitigate this problem by reorganizing class labels based on coarse pre-training results. This process, shown in the right part of Fig.~\ref{fig:low-level-prior idea}, entailed feature visualization by t-SNE~\cite{tsne}, merging similar texture data with different labels, and separating discrete clusters. To further expand our data, we combined an edge-sensitive image matting dataset, resulting in 20,181 samples assigned with 35 reorganized labels. We then fine-tuned the PTPM Net using this restructured data for the final PTPM.
For an intuitive comparison, we show the feature distribution comparison of our PTPM and the image classification-based prior in the supplement. 
The PTPM feature shows better clustering performance than the image classification-based feature, signifying its increased sensitivity to texture changes. 
Additionally, in Fig.~\ref{fig:low-level-prior visualize}, the L2 nearest neighbors of selected samples in different prior spaces show that PTPM's method of measuring texture similarity more closely aligns with human perception.
\begin{table*}[tbp]
	\centering
	\caption{Quantitative comparison (PSNR $\uparrow$, SSIM $\uparrow$) with state-of-the-art BSR methods on $6$ different benchmarks.}
%   \vspace{-0.2cm}
	% \resizebox{\textwidth}{22mm}{
	\setlength{\tabcolsep}{2mm}{
        \begin{tabular}{c| c c| c c | c c | c c | c c | c c}  
			\hline
			\multirow{2}{*}{Method} & \multicolumn{2}{c|}{DIV2K} & \multicolumn{2}{c|}{Urban100} & \multicolumn{2}{c|}{BSDS100} & \multicolumn{2}{c|}{Manga109} & \multicolumn{2}{c|}{Set14} & \multicolumn{2}{c}{Set5} \\
			& PSNR  & SSIM  & PSNR   & SSIM  & PSNR  & SSIM  & PSNR  & SSIM  & PSNR  & SSIM  & PSNR  & SSIM  \\
			\hline
			CDC~(\citeyear{CDC}) & 19.79 &0.4735 & 17.43 & 0.4010 & 20.13 & 0.4384 & 17.64 & 0.5223 & 19.69 & 0.4802 & 18.90 & 0.4717  \\
			DAN~(\citeyear{DAN}) & 20.07 & 0.4577 & 17.74 & 0.4034 & 20.46 & 0.4341 & 18.13 & 0.5229 & 19.83 & 0.4727 & 19.63 & 0.4697  \\
			Real-ESRGAN~(\citeyear{real-esrgan})& 20.08 & 0.5273 & 17.51 & 0.4443 & 20.31 & 0.4383 & 18.76 & 0.6064 & 20.05 & 0.4723 & 19.95 & 0.5125  \\
			SwinIR-GAN~(\citeyear{swinir})& 20.37 & 0.5283 & 17.64 & 0.4562 & 20.15 & 0.4310 & 19.18 & \textbf{0.6251} & 20.21 & 0.4776 & 19.44 & 0.4680  \\
			BSRDM~(\citeyear{BSRDM}) & 20.19 & 0.5330 & 17.18 & 0.4031 & 19.94 & 0.4333 & 17.31 & 0.5437 & 19.17 & 0.4621 & 18.62 & 0.4715 \\
			D2C-SR~(\citeyear{D2CSR}) & 19.44 & 0.4156 & 17.40 & 0.3801 & 20.12 & 0.4199 & 17.54 & 0.4933 & 19.82 & 0.4765 & 18.61 & 0.4229 \\ 
			KXNet~(\citeyear{KXNet}) & 20.10 & 0.4696 & 17.57 & 0.3992 & 20.34 & 0.4341 & 17.84 & 0.5214 & 19.64 & 0.4702 & 19.28 & 0.4650 \\
			MM-RealSR~(\citeyear{MMRealSR}) & 20.60 & \textbf{0.5471} & 17.95 & 0.4585 & 20.34 & 0.4473 & 18.80 & 0.6153 & 20.02 & 0.4817 & 19.84 & \textbf{0.5152} \\ 
			FeMaSR~(\citeyear{femasr})& 20.31 & 0.4918 & 18.01 & 0.4384 & 20.09 & 0.4156 & 19.15 & 0.6024 & 20.18 & 0.4581 & 19.57 & 0.4536  \\
			MRDA~(\citeyear{MRDA}) & 19.91 & 0.4474 & 17.70 & 0.4009 & 20.43 & 0.4328 & 18.07 & 0.5202 & 19.82 & 0.4731 & 19.57 & 0.4666 \\ 
			RTCNet(ours)& \textbf{20.76} & 0.5268 & \textbf{18.40} & \textbf{0.4586} & \textbf{20.91} & \textbf{0.4537} & \textbf{19.52} & 0.6133 & \textbf{20.38} & \textbf{0.4829} & \textbf{20.32} & 0.4931  \\
			\hline
		\end{tabular}
	}
	% }
	\label{tab:sota_cmp}
%  \vspace{-0.2cm}
\end{table*}
\begin{table*}[htbp]
	\centering
	\caption{Perceptual metrics (LPIPS~\cite{lpips} $\downarrow$ ) comparison with state-of-the-art blind super-resolution methods on DIV2K.}
%   \vspace{-0.2cm}
	% \resizebox{\textwidth}{4mm}{
	\setlength{\tabcolsep}{1.7mm}{
		\begin{tabular}{c| c| c | c| c| c| c|c|c|c|c|c}  
			\hline

			Methods & CDC & DAN & SwinIR & Real-ESRGAN & BSRDM & MM-RealSR & KXNet & D2C-SR & FeMaSR & MRDA & RTCNet(ours) \\ 
			\hline
        	LPIPS & 0.7722 & 0.7466 & 0.4739 & 0.5637 & 0.7505 & 0.5724 & 0.7655 & 0.7689 & 0.4480 & 0.7464 & \textbf{0.4390} \\ 
			\hline
		\end{tabular}
	}
	% }
	\label{tab:sota_cmp_per}
%  \vspace{-0.1cm}
\end{table*}

\subsection{Training losses}
\label{sec: loss}
\paragraph{\textbf{Codebook Loss}} This loss optimizes DTPM, including the codebook loss and two correlation constraints:
\begin{equation}
	L_{Codebook} = L_{DTPM} + L_{Rep\ Con} + L_{Rec\ Con}.	
\end{equation}

\paragraph{\textbf{Image Reconstruction Loss}} We use $L1$ and Perceptual Loss~\cite{perceptualloss} as the main reconstruction loss. Following previous work~\cite{QuanTexSR,femasr}, we use a U-net discriminator $D$ in ~\cite{real-esrgan} and a hinge loss as an adversarial loss to get more realistic textures. Given a reconstructed image $I_{Recon}$ and its ground truth image $I_{GT}$, the image reconstruction loss can be formulated as
\begin{equation}
	\begin{aligned}
	\!\!\!\!L_{Rec}(I_{GT},I_{Recon})=&||I_{GT}-I_{Recon}||_1 \\
		 +\lambda_{per} ||\phi_{per}(I_{GT})- &\phi_{per}(I_{Recon})||_1 +\lambda_{adv} E[D(I_{Recon})],
	\end{aligned}
\end{equation}
where $\phi_{per}$ denotes a pre-trained VGG16~\cite{vgg} network.

\paragraph{\textbf{PTPM Loss}}
We integrate the PTPM prior into the DTPM's training by applying scale-matched texture prior regularization. Specifically, the global texture priors are the activations of the 5th ReLU of the ImageNet-pretrained VGG19~\cite{vgg} network $\phi_{img}$ and the local-friendly priors are the activations of the 2nd Max Pooling of our PTPM Net $\phi_{p}$. We extract the texture priors from the HR images. The PTPM supervision $L_{PTPM}$ is computed between the quantized features $\widehat{F}$ and the corresponding texture priors which can be formulated as
\begin{equation}
	\begin{aligned}
		L_{PTPM}(I_{HR}, \widehat{F}) = ||\phi_{img}(I_{HR}) - \phi_{p_g}(\widehat{F_g})||_2 + ||\phi_{p}(I_{HR}) - \phi_{p_l}(\widehat{F_l})||_2,
	\end{aligned}
\end{equation}
where $\phi_{p_{l,g}}$ are single convolution layer to transfer from the codebook space to the prior space.
The total PTPM supervision is the sum of the supervision on the quantized features of two scales as
\begin{equation}
		L_{PTPM} = L_{PTPM}(I_{HR},\widehat{F_{HR}}) + L_{PTPM}(I_{HR},\widehat{F_{LR}}).
\end{equation}

\paragraph{\textbf{Overall Loss}}
The overall loss is then defined as
% \vspace{-0.2cm}
\begin{equation}
	L_{total}=L_{Codebook}+ L_{PTPM}.
\end{equation}

% !TEX root = sample-xelatex.tex
\section{Experiments}
\label{sec:experiments}
\subsection{Datasets and Evaluation Metrics}
\paragraph{\textbf{Prior Pre-training Datasets}}
Our coarse patch classification dataset is based on the ADE20K~\cite{ADE20K} semantic segmentation dataset and expanded using the SIM~\cite{SIM} image matting dataset. Following the strategy in Sec~\ref{Sec:low-level prior}, we generate a final dataset with $17,880$ training samples and $2,301$ validation samples.

% \vspace{-0.1in}
\paragraph{\textbf{Super-Resolution Training Dataset}} We build an overall training dataset including DV2K~\cite{DIV2K}, DIV8K~\cite{DIV8K}, Flickr2K~\cite{Flickr2K}, and OST~\cite{wang2018sftgan} datasets. HR patches are generated using the following approach: \textit{1)} crop large images into non-overlapping $512\times512$ patches; \textit{2)} apply the blur detection method~\cite{kim2018defocus} to each patch to filter out blurred patches with a blurred area greater than $95\%$. Our final training dataset contains $123,395$ HR patches, while we generated LR patches for each iteration using the widely used degradation model proposed in~\cite{BSRGAN}.

% \vspace{-0.1in}
\paragraph{\textbf{Super-Resolution Test Datasets}} We evaluated the performance of our model using six benchmark datasets, namely DIV2K~\cite{DIV2K}, Set14~\cite{Set14}, Set5~\cite{Set5}, Urban100~\cite{urban100}, BSD100~\cite{BSDS100}, and Manga109~\cite{manga109}. We used the mixed degradation model described in~\cite{real-esrgan} and ~\cite{BSRGAN} for LR generation.
The diverse datasets with complex degradation, allowed for a comprehensive performance evaluation. A 4x downsampling was used for all experiments.
% \vspace{-0.1in}
\paragraph{\textbf{Evaluation Metrics}} We used Peak Signal to Noise Ratio (PSNR) and Structural Similarity (SSIM) to evaluate the quality of generated images. In addition, for better perceptual evaluation, we also use the Learned Perceptual Image Patch Similarity (LPIPS)~\cite{lpips}.
% \vspace{-0.2in}
\paragraph{\textbf{Implementation Details}} We implement our model using the PyTorch framework. In both low-level prior pretraining and SR training, we use an Adam~\cite{adam} optimizer with $\beta_1=0.9, \beta_2= 0.99, lr=1\times 10^{-4}$. The number of codes in both scale codebooks is set to $512$. The RTCNet is trained with a batch size of $16$ and a HR patch size of $256\times 256$ on 4 NVIDIA V100 GPUs for about 4 days.

\begin{figure*}[bthp]
	%\vspace{-0.4cm}
	\centering
		\includegraphics[width=1\textwidth]{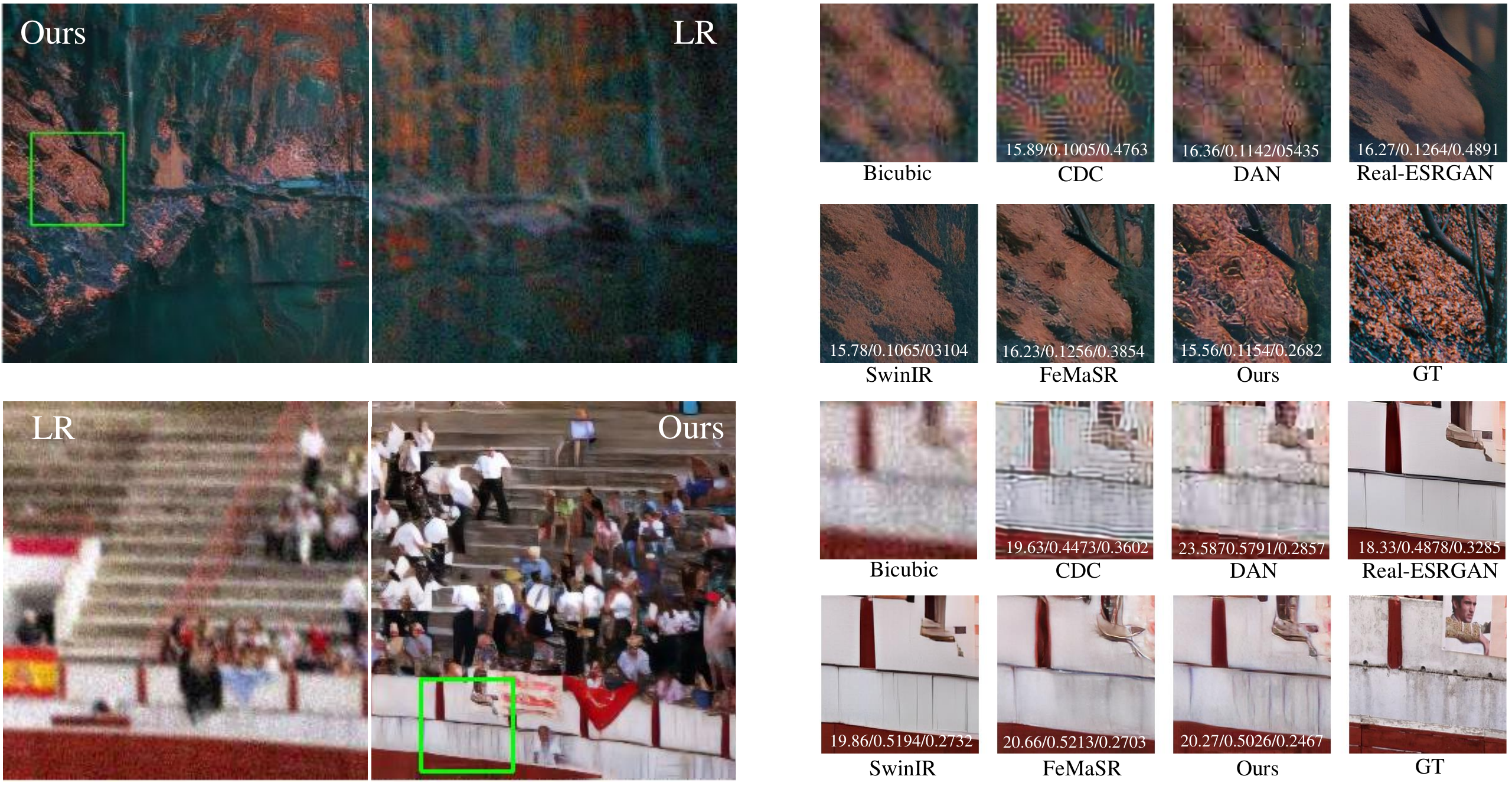}
		% \includegraphics[width=0.92\textwidth]{./sota_cmp.pdf}
		%  \vspace{-0.1in}
		\caption{Visual comparison with other blind super-resolution images. The PSNR/SSIM/LPIPS values are shown at the bottom of the images. The captions in the images below have the same meaning as the descriptions provided here.}
		\label{fig:sota_cmp}
		% \vspace{-0.1in}
	\end{figure*}
% \vspace{-0.1cm}
\subsection{Comparison with SOTA}

We compared the proposed RTCNet with $10$ recent state-of-the-art blind SR methods, including CDC~\cite{CDC}, DAN~\cite{DAN}, Real-ESRGAN~\cite{real-esrgan}, SwinIR-GAN~\cite{swinir}, BSRDM~\cite{BSRDM}, D2C-SR~\cite{D2CSR}, KXNet~\cite{KXNet},  MM-RealSR~\cite{MMRealSR}, FeMaSR~\cite{femasr} and MRDA~\cite{MRDA}. We compare our method with these approaches using the published codes and weights from the official public GitHub repos.

As shown in Tab.~\ref{tab:sota_cmp}, our method achieves the best PSNR/SSIM performance on almost all $6$ datasets. 
In Fig.~\ref{fig:sota_cmp}, we compare the restored images of different BSR methods. First, consistent with the results in Tab.~\ref{tab:sota_cmp} and Tab.~\ref*{tab:sota_cmp_per}, DAN and CDC have limited recovery effects for complex degraded images. Second, Real-ESRGAN and SwinIR-GAN tend to confuse noise and texture. They erase texture details as noise and cause over-smoothing problems. Besides, FeMaSR mistakes some noise for texture, resulting in noisy texture generation.
On the contrary, since our DTPM effectively maintains the cross-resolution consistency of texture codebooks, it is more robust to low-resolution degradation. It can reasonably distinguish between texture and degradation, ensuring the restoration of realistic textures while denoising. The multi-scale structure and low-level friendly priors further improve the restoration of local fine textures. In general, our RTCNet achieves state-of-the-art performance in quantitative metrics and human perception.

% \vspace{-0.1in}
\subsection{Ablation Study}

% \vspace{-0.1in}
\paragraph{\textbf{Effectiveness of Cross-Resolution Correlation.}} To verify the effectiveness of cross-resolution constraints, we conduct an ablation study on the two cross-resolution strategies used: cross-resolution representation consistency (Rep. C.) and cross-resolution reconstruction consistency (Rec. C.). As shown in Tab.~\ref{Tab: DTPM_ablation2}, both of them can effectively improve the performance of DTPM. This is because the cross-resolution constraint forces the LR representation in the codebook to be closer to the HR, making it as distinguishable as the HR in the codebook space. And the combination of the two can further enhance the improvement.

\begin{table}[t]
	\centering
    \caption{Ablation of DTPM. Rows 1-4: Ablation of DTPM consistency constraints. Rows 5-7: Ablation of components in hierarchical codebook learning. Rep. C.: Representation Consistency; Rec. C.: Reconstruction Consistency. H.S.: Hierarchical Structure; D2S: Deep-to-Shallow strategies.} 
	% \vspace{-0.1in}
	\setlength{\tabcolsep}{2.5mm}{
	\begin{tabular}{c c | c c c | c c}  
		\hline
		%	global prior & 
		Rep. C. & Rec. C. & H.S. &D2S & PTPM & PSNR & SSIM \\
		\hline
		%	ImgNet Prior &
		$\times$ & $\times$ & $\times$ & - & - & 19.90 & 0.5004\\
		%	ImgNet Prior &
		$\times$ & $\checkmark$ & $\times$ & - & - & 20.17& 0.4974\\
		$\checkmark$ & $\times$ & $\times$ & - & - & 20.19& 0.4990\\
		$\checkmark$ & $\checkmark$ & $\times$ & - & - & 20.59 & 0.5180\\
		\hline
		$\checkmark$ & $\checkmark$ & $\checkmark$ & $\times$ & $\times$ & 19.53 & 0.4781 \\
		$\checkmark$ & $\checkmark$ & $\checkmark$ & $\checkmark$ & $\times$ & 20.52& 0.5215\\
		%	ImgNet Prior &
		$\checkmark$ & $\checkmark$ & $\checkmark$ & $\checkmark$ & $\checkmark$ & \textbf{20.76} & \textbf{0.5268}\\
		\hline
	\end{tabular}
	}
	\label{Tab: DTPM_ablation2}
	% \vspace{-0.2in}
\end{table}
% \vspace{-0.1in}
\paragraph{\textbf{Effectiveness of Hierarchical Structure.}}
We validated the effectiveness of prior feature regularization and deep-to-shallow training strategy for multi-scale codebook training in Tab.~\ref{Tab: DTPM_ablation2}. Training a multi-scale model from scratch leads to insufficient texture learning due to the more diverse and sensitive texture degradation at the local scale, making its performance even worse than that of the single-scale model (see rows 4 and 5, Tab.~\ref{Tab: DTPM_ablation2}). The addition of the deep-to-shallow training strategy stabilizes the learning of the local scale codebook by the well-trained encoders and global features, which significantly improves the restoration of large textures. Notably, the performance of the deep-to-shallow strategy is not significantly better than the single-scale model, while after adding the PTPM regularization on this basis, the result is better than the single-scale model. This shows that the full multi-scale model can achieve better performance for fine textures than the single-scale model through hierarchical texture learning, but the learning of local-scale texture is challenging and requires the assistance of a low-level texture-friendly prior.

\begin{table}[t]
	\centering
% \vspace{-0.1cm}
    \caption{Comparison of DTPM with the high-resolution reconstruction-based codebook of FeMaSR~\cite{femasr} in the DIV2K validation set. The '*' indicates that we conduct a single-scale codebook without the hierarchical structure.}
    % \vspace{-0.3cm}
    \setlength{\tabcolsep}{3.5mm}{
	\begin{tabular}{c| c c | c}  
		\hline
		& PSNR & SSIM & Codebook Use ratio\\
		\hline
		FeMaSR~\cite{femasr} & 20.31 & 0.4918 & 33 / 1024 \\
		DTPM*& \textbf{20.59} & \textbf{0.5180} & 396 / 512\\
		\hline
	\end{tabular}
	}
	% \vspace{-0.2in}
	\label{Tab: DTPM_ablation}
\end{table}
\begin{figure}[t]
		\centering
		% \vspace{-0.1in}
		\includegraphics[width=0.95\columnwidth]{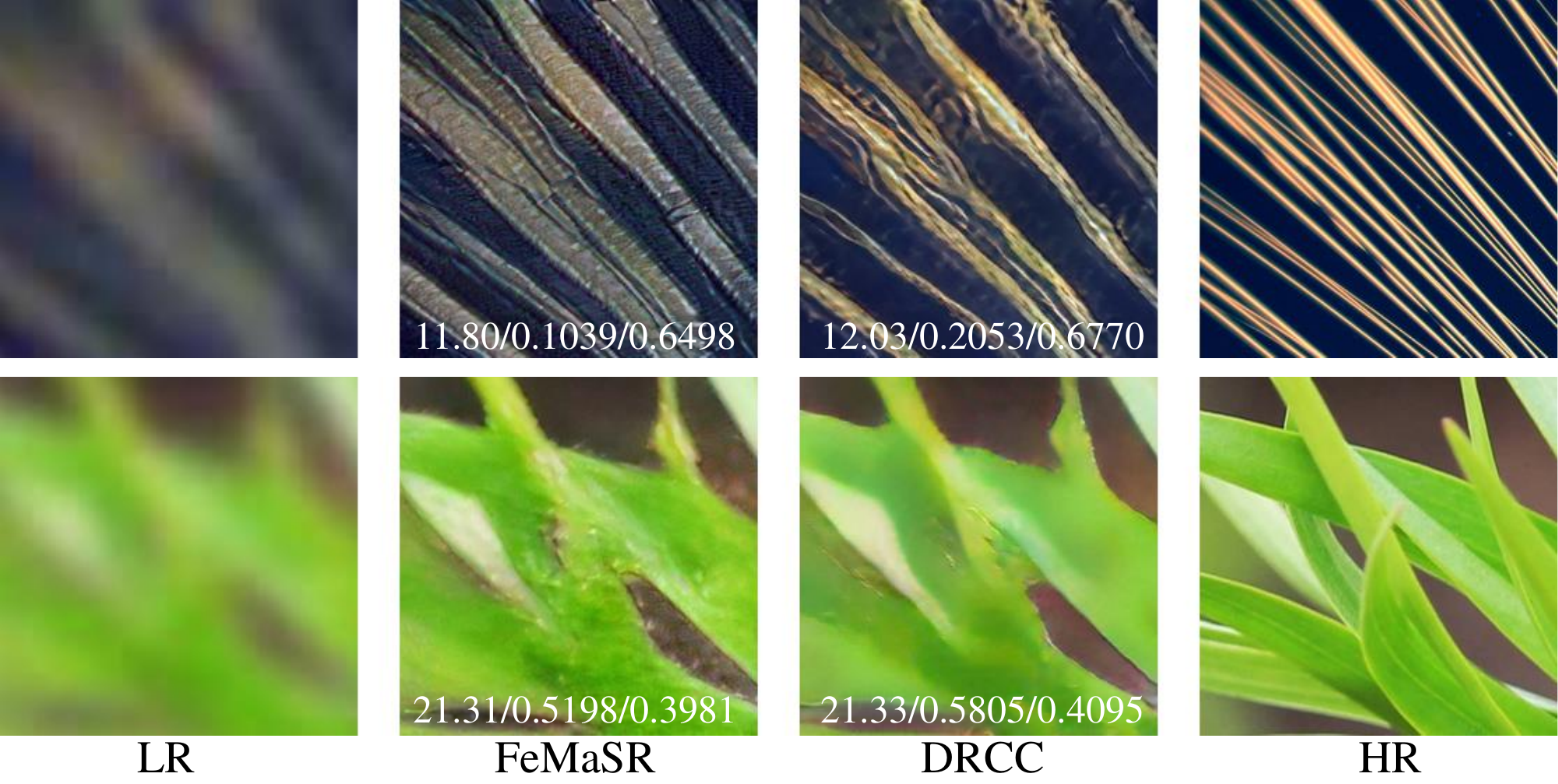}
		\vspace{-0.4cm}
		\caption{Visual comparison between single-scale DTPM and FeMaSR~\cite{femasr}. }
		\label{fig: DTPM_cmp}
	%  \vspace{-0.2in}
	\end{figure}

% \vspace{-0.1in}
\paragraph{\noindent\textbf{Comparison with Previous Codebooks.}}  To verify the superiority of our proposed DTPM, we compare it with the high-resolution reconstruction-based codebook of FeMaSR~\cite{femasr}. For fairness, both of them used the bottleneck model structure (codebook at x8 downsampling) and trained with our overall loss except for the local-scale PTPM loss. As shown in Tab.~\ref{Tab: DTPM_ablation} and Fig.~\ref{fig: DTPM_cmp}, DTPM outperforms FeMaSR~\cite{femasr} in both quantitative and qualitative results. Compared to FeMaSR~\cite{femasr}, our single-scale DTPM has a more stable and realistic texture generation in heavy degradation. To show the advantage of DTPM more intuitively, we also statistically analyze the codebook utilization in the inference stage. As shown in Tab.~\ref{Tab: DTPM_ablation}, only $3.2\%$ of FeMaSR's codebook was used during inference. Such an over collapse indicates the inadequacy and indistinguishability of the HR-reconstruction-based codebook when applied to LR data. This limitation of the codebook limits the variety of textures that can be generated, resulting in unrealistic recovery during super-resolution. In contrast, DTPM has a wider range of codebook usage. This is because the codebook space is trained with LR data and contains more discriminative LR representations under the cross-resolution consistency constraint. Benefiting from this, DTPM achieves more diverse texture generation and stronger stability (less noise on row 2 and clear texture on row 1 in Fig.~\ref{fig: DTPM_cmp}).

\begin{table}[t]
	% \vspace{-0.1in}
    \caption{Comparison of different priors used to learn the local-scale codebook in the DIV2K validation set. ($\dag$ denotes the coarse prior before label refinement and fine-tuning).}
    % \vspace{-0.3cm}
	\centering
	\setlength{\tabcolsep}{5.5mm}{
	\begin{tabular}{c| c c}  
	\hline
	 Local Prior &  PSNR &  SSIM \\
	\hline
	 - &  20.52 &  0.5215\\
	  ImgNet-Classification Prior &   20.57 &  0.5240\\
	  Patch-aware Texture Prior$\dag$ &   20.67 &  \textbf{0.5314}\\
	  Patch-aware Texture Prior &  \textbf{20.76} &  0.5268 \\
	\hline
	\end{tabular}
	}
\label{Tab:prior_cmp}
% \vspace{-0.2cm}
\end{table}
\begin{figure}[t]
% \vspace{-0.2cm}
	\includegraphics[width=0.96\columnwidth]{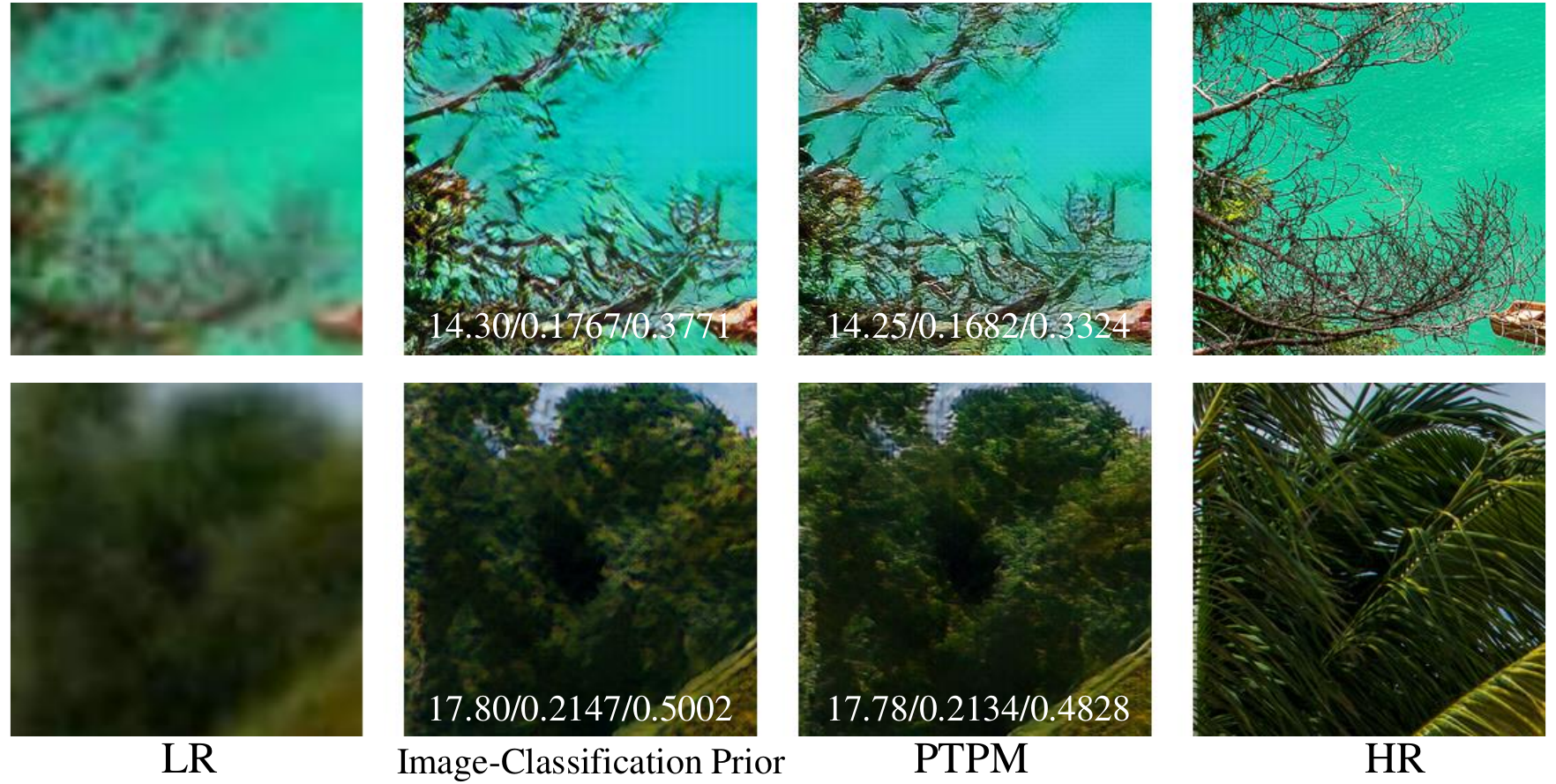}
	% \includegraphics[width=0.9\columnwidth]{./low-level-prior3.pdf}
	% \vspace{-0.1in}
	\caption{Visual comparison of local-scale codebooks trained with different prior features.}
	\label{fig:low-level-prior}
	% \vspace{-0.2in}
\end{figure}

% \vspace{-0.1in}
\paragraph{\textbf{Effectiveness of PTPM in blind super-resolution}}
To verify the superiority of our PTPM for low-level texture learning, we compared the impact of different semantic features on the learning of the local codebook in RTCNet. As shown at the bottom of Fig.~\ref{fig:low-level-prior} and in Tab.~\ref{Tab:prior_cmp}, while all pre-trained priors improve the texture restoration performance, our PTPM prior outperforms the ImageNet-based pre-trained prior. This superiority can be attributed to our PTPM's better perception of low-level texture correlations. In addition, patch-level pre-training and texture-oriented label organization both improve the performance of the PTPM.
To demonstrate the superior ability of the PTPM prior to distinguish different types of textures compared to the image classification-based prior, we analyzed the frequency distribution of different codes used for super-resolution on the OST dataset in the Supplement. As can be observed, compared to the ImageNet Classification pre-trained priors, PTPM shows more distribution differences between the "grass" and "plant" categories, which have more overlapping semantic labels, and has a smaller difference in the "sky" and "water" categories, which have different semantic categories but relatively similar textures. This shows that our pre-training strategy enables PTPMNet to pay more attention to the correlation of local texture information by excluding high-level information from the pre-training.

% !TEX root = sample-xelatex.tex
% \vspace{-0.2in}

\subsection{Limitation and Discussion}
First, 
	by observing the results, we find that RTCNet has some limitations when dealing with regular texture restoration, especially for data types that have plenty of such textures, such as buildings (examples in Supplement). This problem also occurs with the previous codebook-based method, which we will investigate in future work.
Second, 
	we find that the improvement of RTCNet is more obvious in the heavily degraded samples than in the lightly degraded samples (perhaps no improvement in some light samples) (Fig.~\ref{fig:dtpm_femasr_cmp}.c). We speculate that the notable improvements in the heavily degraded data are due to increased matching confusion, a scenario where RTCNet performs optimally. Conversely, light degradation with less confusion can also be handled by previous methods, leading to marginal improvements. Although both data types are common in applications, we argue that the correction of complex degraded data has great challenges and value for super-resolution (SR) tasks.
Third, 
	the improvements brought by PTPM are not very considerable and stable, indicating that larger valid datasets and a more refined pretraining strategy is valuable for better performance.
Furthermore, 
	based on experience, pre-training features tend to adapt more effectively to data with strong domain priors, suggesting that applying the codebook method in combination with pre-training strategies to specific types of data may be a direction worth investigating.

\section{Conclusion}
\label{sec:conclusion}
% \vspace{-0.2cm}
In this paper, we have presented the Rich Texture-aware Codebook Network (RTCNet) framework for blind image super-resolution. With our proposed Degradation-aware Texture Codebook Module, we allow for more efficient modeling of LR-HR correspondences than previous single HR reconstruction pre-training. The architectures of DTPM allow it to model large and fine textures separately. In addition, we build the low-level friendly Patch-aware Texture Prior Module (PTPM) which further improves the performance of DTPM. Various experiments on different benchmarks show that our RTCNet achieves state-of-the-art performance.
% \section*{Acknowledgement}
% This work was supported by the National Natural Science Foundation of China under Grant 62072271.
\nocite{*}
\bibliographystyle{ACM-Reference-Format}
\balance
\bibliography{11_references}
\newpage
% !TEX root = sample-xelatex.tex
\appendix
\section{Supplementary materials}

\label{sec:appendix}
%\subsection{test}

\subsection{LR Confusion in BSR data} 
\begin{figure}[htbp]
	\centering
\includegraphics[width=\linewidth]{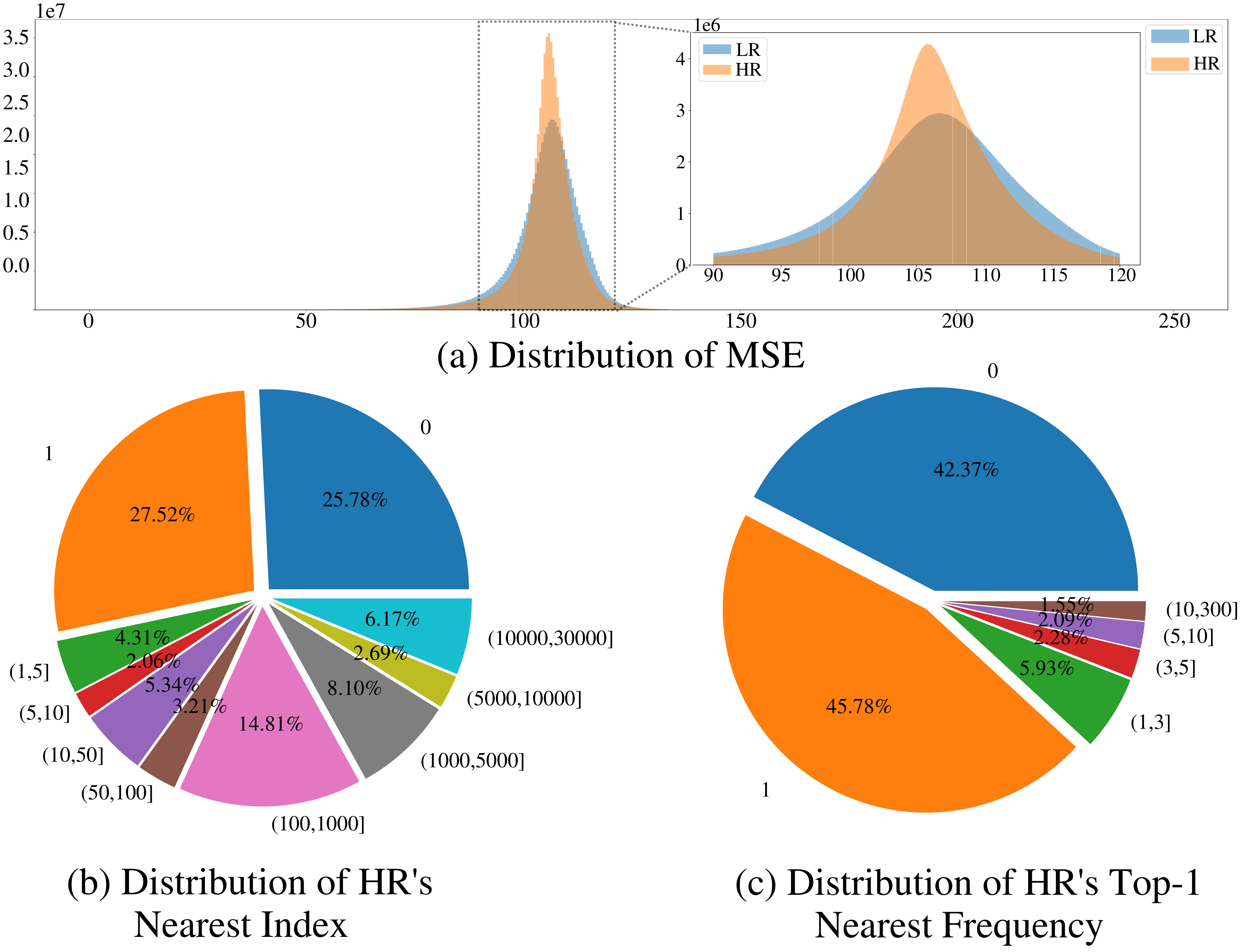}
\caption{The MSE statistics of LR-HR data in validation datasets.}
\label{fig:mse_dis}
\end{figure}

This section presents a statistical analysis of the LR data from all the validation datasets utilized to show the universal confusion phenomenon observed in LR data compared to HR data. We densely cropped all HR images and their corresponding LR versions, which have the same size as the HR version after bicubic upsampling, into $128\times 128$ patches (a total of 26,753 patches). We then computed the mean squared error (MSE) between all HR and LR patches. First, we analyzed the distributions of both HR and LR patches using the MSE, as shown in Figure~\ref{fig:mse_dis}.a. The plot highlights that HR patches have a more concentrated MSE distribution, while LR patches have a more dispersed one. This indicates that the LR data are more prone to confusion. Second, we extracted the index of HR patches in the nearest HR patch sorting of its LR patch. As shown in Figure~\ref{fig:mse_dis}.b, the index is dispersed, with a significant proportion not being the top-1 nearest. This implies that many LR patches have a closer MSE distance to other HR patches than their corresponding HR patches. Furthermore, Fig. \ref{fig:mse_dis}.c shows the frequency of selection of each HR patch as the nearest one to different LR patches. The figure shows that there is a large partition of non-1 frequency, indicating a large part of the LR-HR mismatch. Although the MSE statistic is not entirely suitable for evaluating the similarity between patches, the considerable partition of mismatches between HR patches and their LR counterparts suggests the confusion caused by blind degradation and the complex correlation it introduces.

\subsection{Detailed Comparison between DTPM and FeMaSR}
\begin{figure}[htbp]
	% \vspace{-0.1in}
	\centering
	\includegraphics[width=0.95\linewidth]{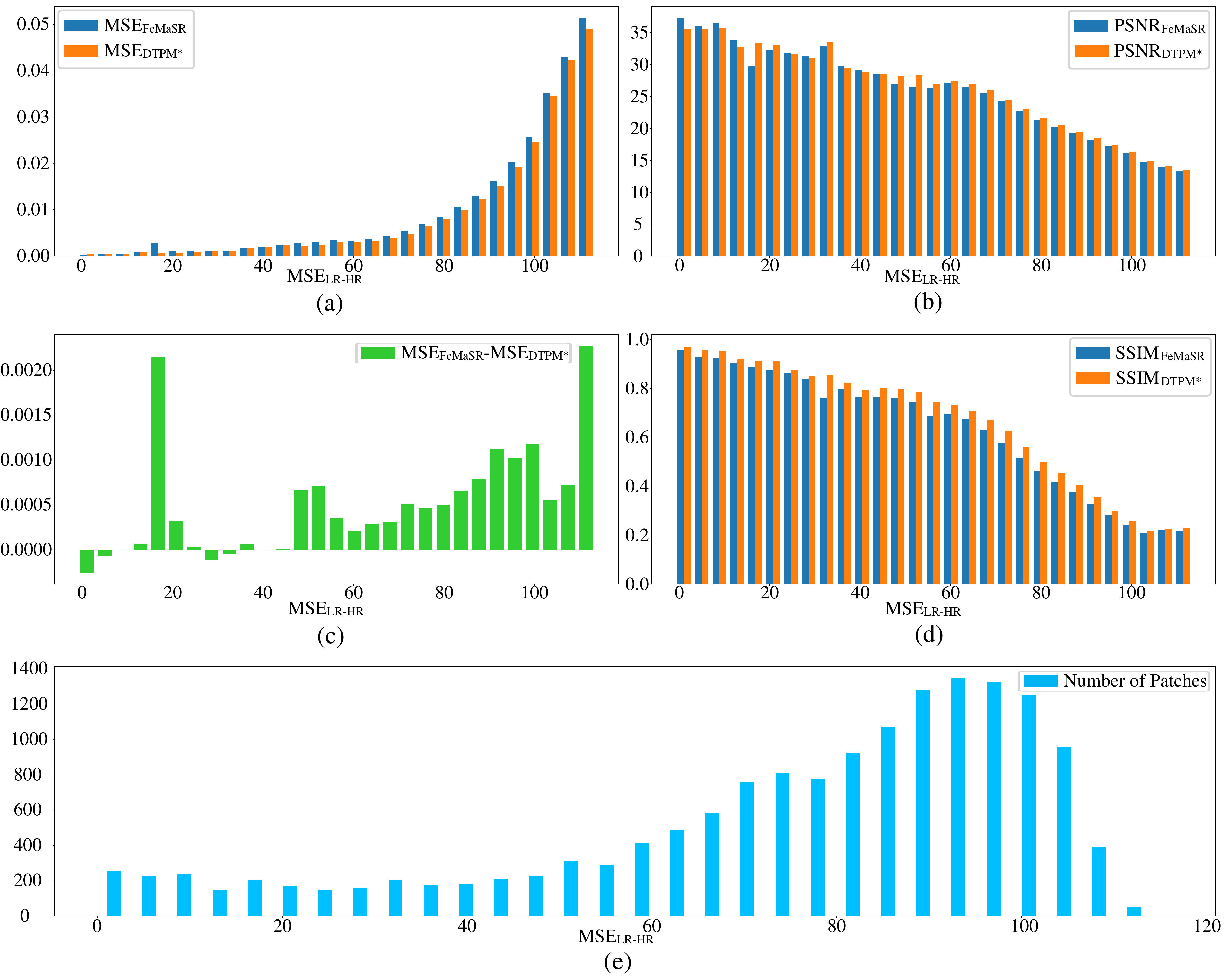}
	\vspace{-0.2in}
	\caption{Detailed comparison between DTPM and FeMaSR. (a) Distribution of MSE between LR and HR. (b) Distribution of PSNR between SR and HR. (c) Distribution of PSNR gain of DTPM over FeMaSR. (d) Distribution of SSIM between SR and HR. (e) Number distribution of image patches with different LR-HR MSE.}
	\label{fig:dtpm_femasr_cmp}
\end{figure}

In this section, we performed a comparative analysis of our DTPM and previous high-resolution reconstruction-based codebooks (using FeMaSR as an example). We statistically investigated the performance improvements of DTPM over the FeMaSR method on samples of varying difficulty in the DIV2K validation set. Specifically, we divided the high-resolution (HR), low-resolution (LR), and their super-resolution (SR) results into $128\times 128$ patches(15585 in total). We used the mean squared error (MSE) distance between LR and HR as a simple measure of sample difficulty, with smaller values indicating easier samples and larger values indicating more difficult samples. We compared and plotted the measurements including MSE(Fig.~\ref{fig:dtpm_femasr_cmp}.a), Peak Signal to Noise Ratio (PSNR, Fig.~\ref{fig:dtpm_femasr_cmp}.b), and Structural Similarity Index (SSIM, Fig.~\ref{fig:dtpm_femasr_cmp}.d) of the SRs of DTPM and FeMaSR under different levels of difficulty. To better illustrate the advantages of DTPM on difficult samples, we also investigate the performance gain of DTPM over the FeMaSR method for different sample difficulty levels(Fig.~\ref{fig:dtpm_femasr_cmp}.c). As shown in Fig.~\ref{fig:dtpm_femasr_cmp}, compared to FeMaSR, our DTPM has achieved improvements in different levels of difficulty, especially for samples with higher difficulty Tab.~\ref{fig:dtpm_femasr_cmp}.c. This verifies the good adaptability of DTPM to LR data, and thanks to its mining of texture cross-resolution consistency, DTPM can better distinguish different types of textures and perform diverse reconstructions for more difficult samples.

\subsection{Validation of Hierarchical texture learning of multi-scale structure}
\begin{figure*}[htbp]
	\centering
	\includegraphics[width=0.85\linewidth]{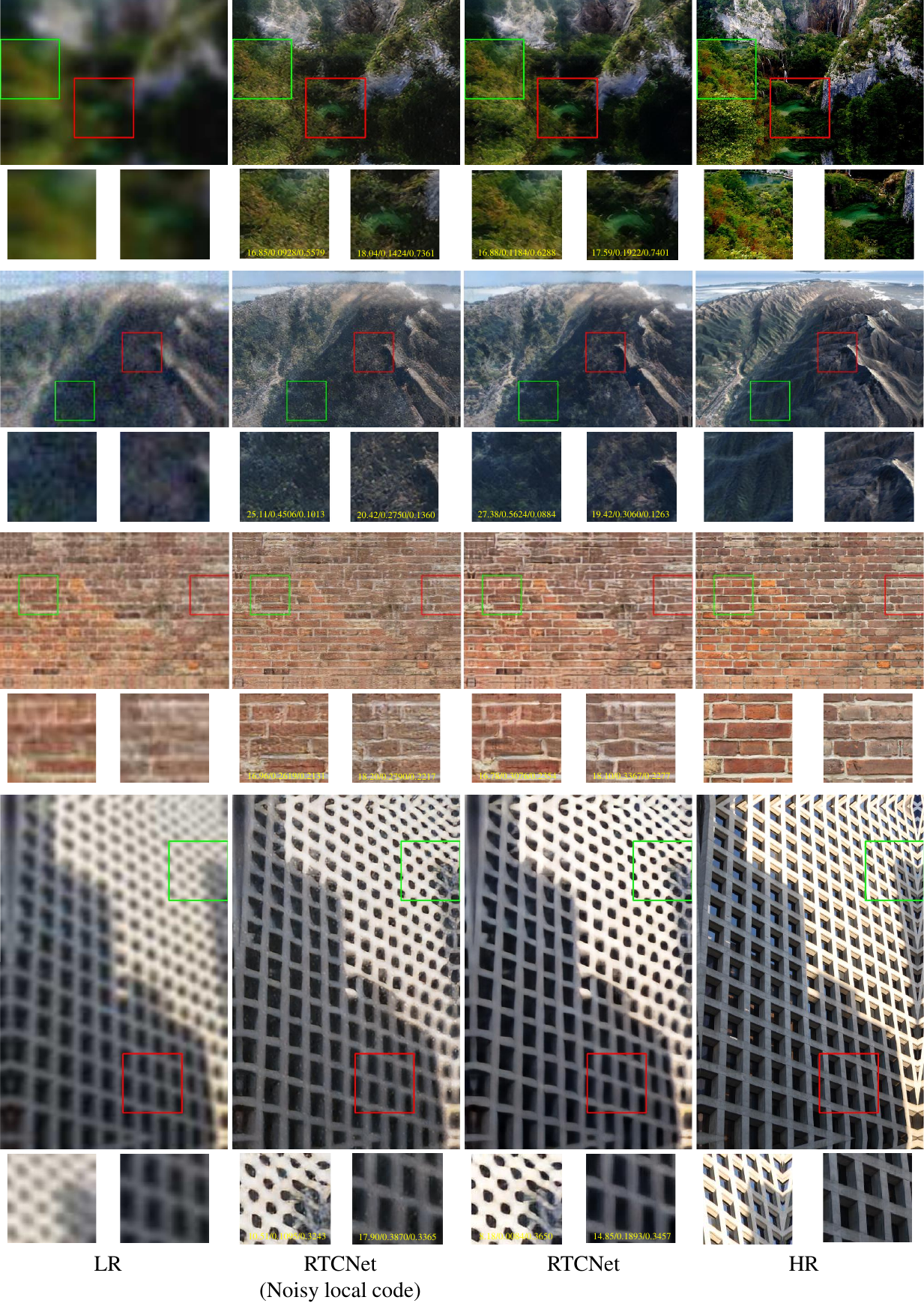}
	\caption{Reconstruction Comparison of the local-scale quantized features generated by random noise and the matching local-scale quantized features obtained from the input.}
	\label{fig:ms_noise_cmp}
\end{figure*}

To better understand the advantage of hierarchical codebooks for texture learning, We explore the texture content learned at different scales in the hierarchical codebook architecture. Specifically, during the quantization process of local-scale DTPM, we replace the quantized features obtained from the low-resolution input with the noise features generated by the random indexes, thereby erasing the influence of the local scale feature during the reconstruction process. Quantitative and qualitative results are shown in Tab.~\ref{Tab: ms_noise_cmp} and Fig.~\ref{fig:ms_noise_cmp}, respectively.

\begin{table}[htbp]
	\caption{Comparison between full hierarchical structure DTPM and noisy-local DTPM.}
	\centering
	\setlength{\tabcolsep}{5mm}{
	\begin{tabular}{c | c c}  
		\hline
		Method & PSNR & SSIM \\
		\hline
		RTCNet(noisy local-scale code) & 20.17  & 0.4928 \\
		RTCNet  & \textbf{20.76} & \textbf{0.5268}\\
		\hline
	\end{tabular}
	}
	\label{Tab: ms_noise_cmp}
\end{table}
 In Fig.~\ref{fig:ms_noise_cmp},  when the local-scale information is missing, the detailed texture restoration is heavily affected, causing unrealistic fined texture reconstruction. In contrast, the global contour and large-scale textures are not significantly affected. This shows that the hierarchical structure learns textures of different sizes with different-scale codebooks. Global codebook and local codebook are responsible for global- and local-scale textures separately. Such a strategy improves the model's ability to model different textures and increases the diversity of textures in the reconstruction results.

\subsection{More Analysis Experiment of PTPM Prior Features}
\begin{figure*}[htbp]
	\centering
\includegraphics[width=\linewidth]{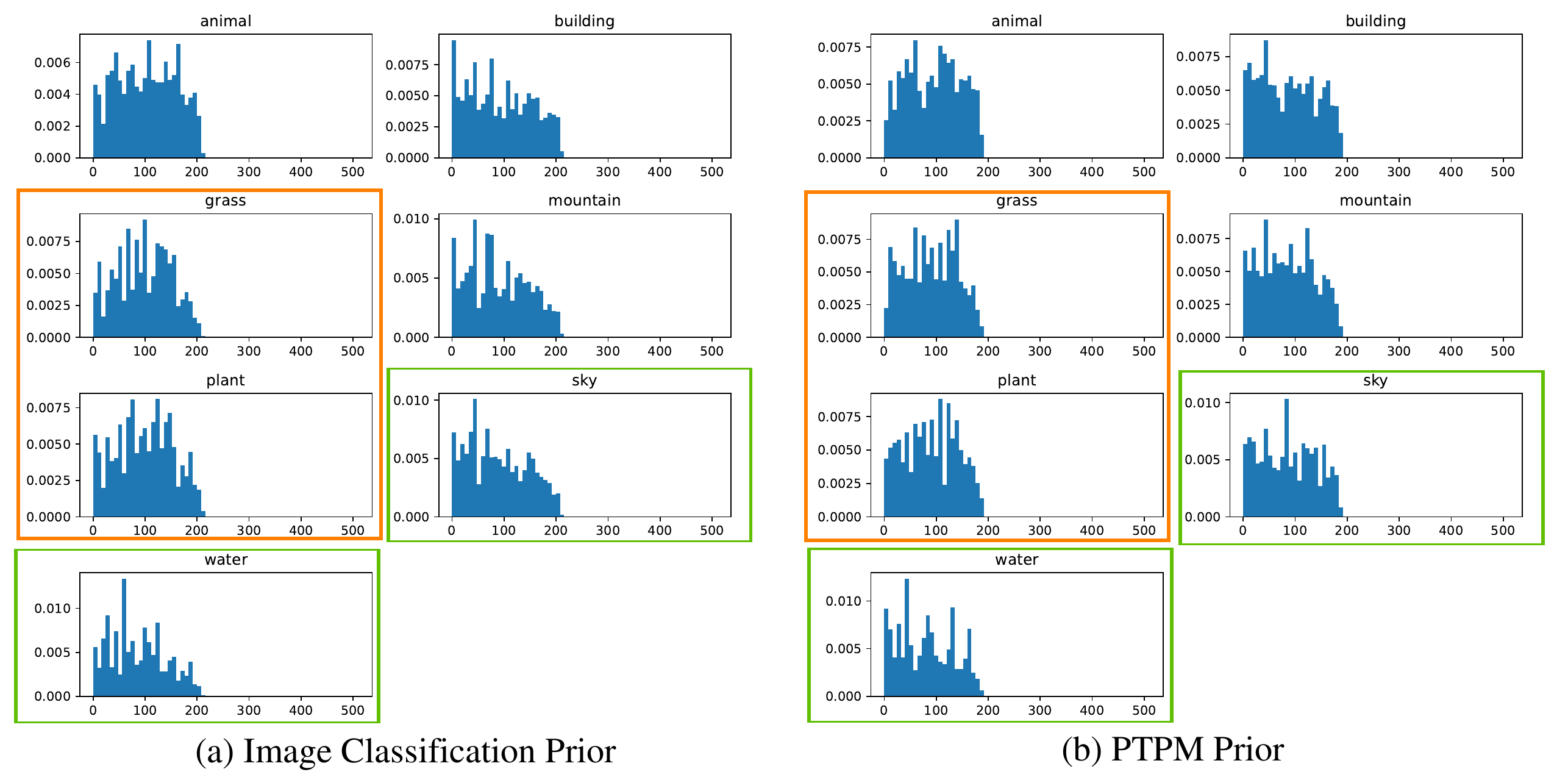}
\caption{The frequency distribution of different codes used during super-resolution on the OST dataset.}
\label{fig:ost_dis}
\end{figure*}

\begin{figure*}[htbp]
		\centering
	\includegraphics[width=\linewidth]{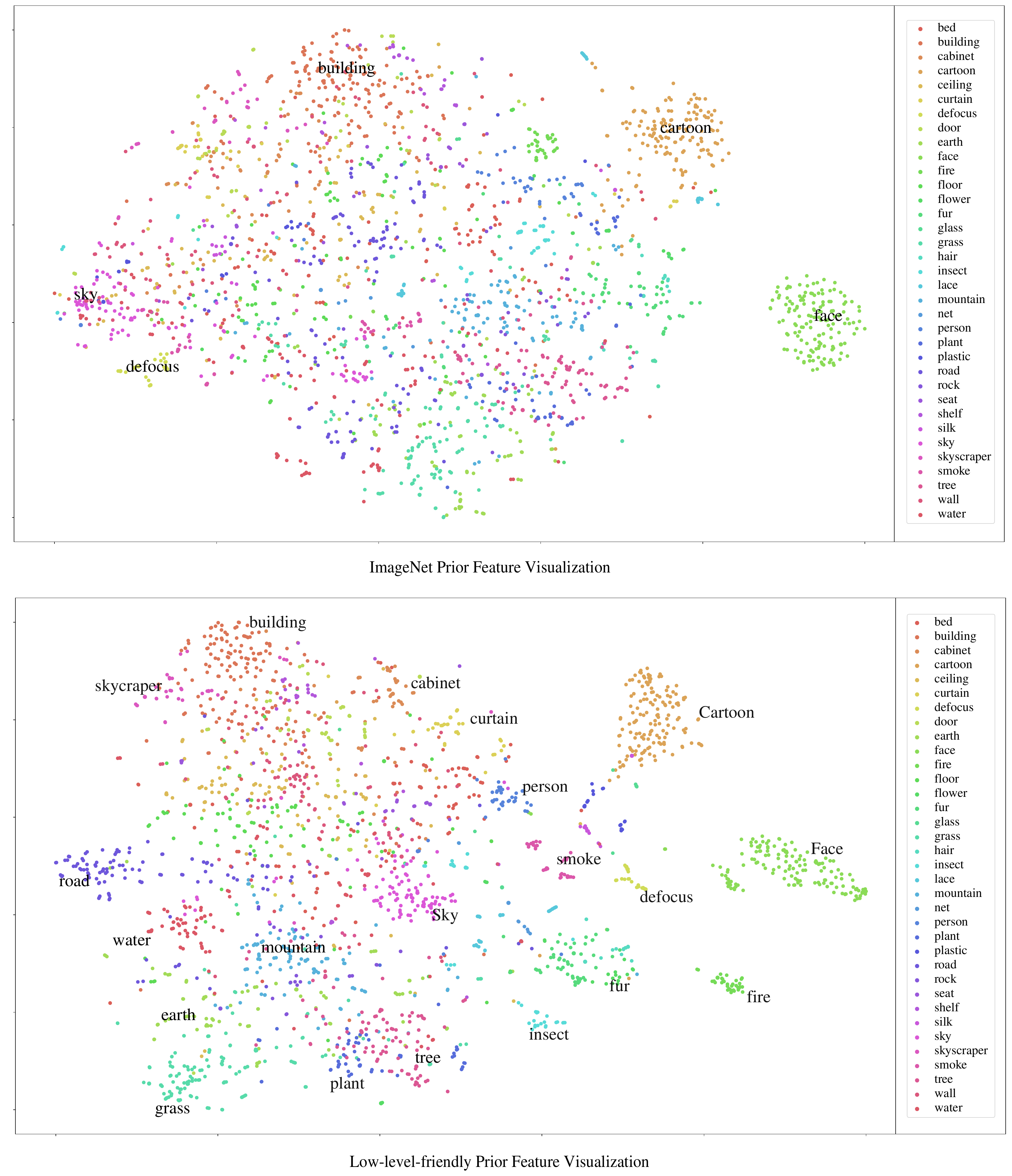}
	\caption{The detailed t-SNE visualization of different prior features extracted from the images of our low-level patch classification validation dataset with the legend of patch-classification dataset classes.}
	\label{fig:tsne_cmp2}
\end{figure*}

We present the detailed visualized comparison using t-SNE dimensionality reduction between image-classification-based priors and our PTPM priors in Fig.~\ref{fig:tsne_cmp2}. Compared with image-classification-based prior features, our PTPM priors have better clustering performance, meaning more sensitivity to local texture similarity. To further illustrate the difference between our PTPM prior and the ImageNet prior in the process of learning low-level texture, we conducted super-resolution statistics on the OST dataset. The high-resolution images in the OST dataset were divided into seven categories according to rough textures, including animal, building, plant, grass, sky, water, and mountain. We degraded the HR data in the OST dataset and perform BSR on them. Then we counted the using frequency of each code in the codebook during the super-resolution process by category. By comparing the distribution of code used when facing different textures, we compare the learned code spaces' rationality on texture perception. As can be observed in Fig.~\ref{fig:ost_dis}, when compared to the ImageNet-Classification-pretrained priors, PTPM exhibits more distribution differences between the “grass” and “plant” categories, which have more overlapped semantic labels, and has a smaller difference in the “sky” and “water” categories, which have different semantic categories but relatively similar textures. This shows that our pre-training strategy enables PTPMNet to pay more attention to the correlation of local texture information by excluding high-level information from the pre-training.

\subsection{Detailed Structure of RTCNet.}
\begin{figure*}[htbp]
	\centering
	\includegraphics[width=\linewidth]{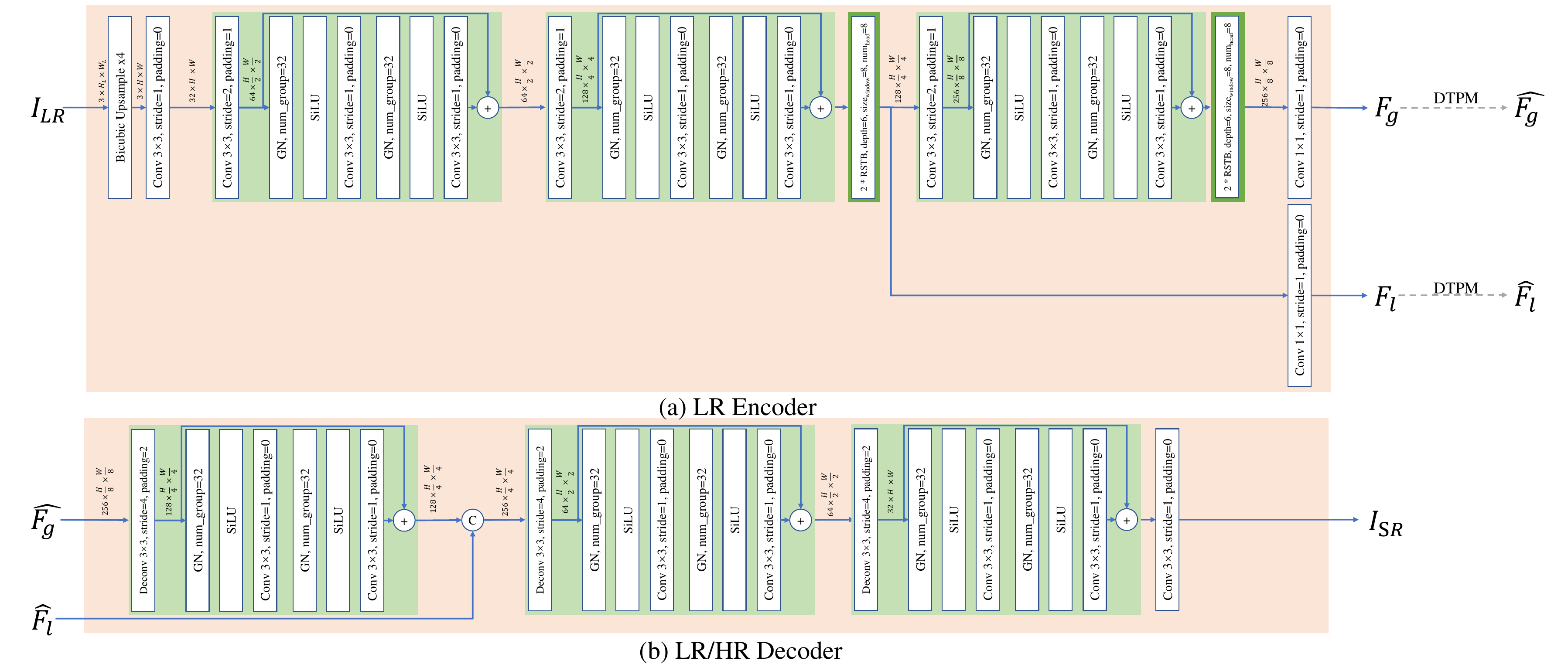}
	\caption{Detailed Structure of the encoders and decoders of RTCNet. The HR encoder is the same as the LR encoder but without the bicubic upsampling layer and RTSBs.}
	\label{fig:model_hyper}
\end{figure*}

In complementary to the RTCNet framework in the paper, we provide the details of hyperparameters for the encoders and decoders in Fig.~\ref{fig:model_hyper}.

\subsection{More Results}
\begin{figure*}[htbp]
		\centering
	\includegraphics[width=\linewidth]{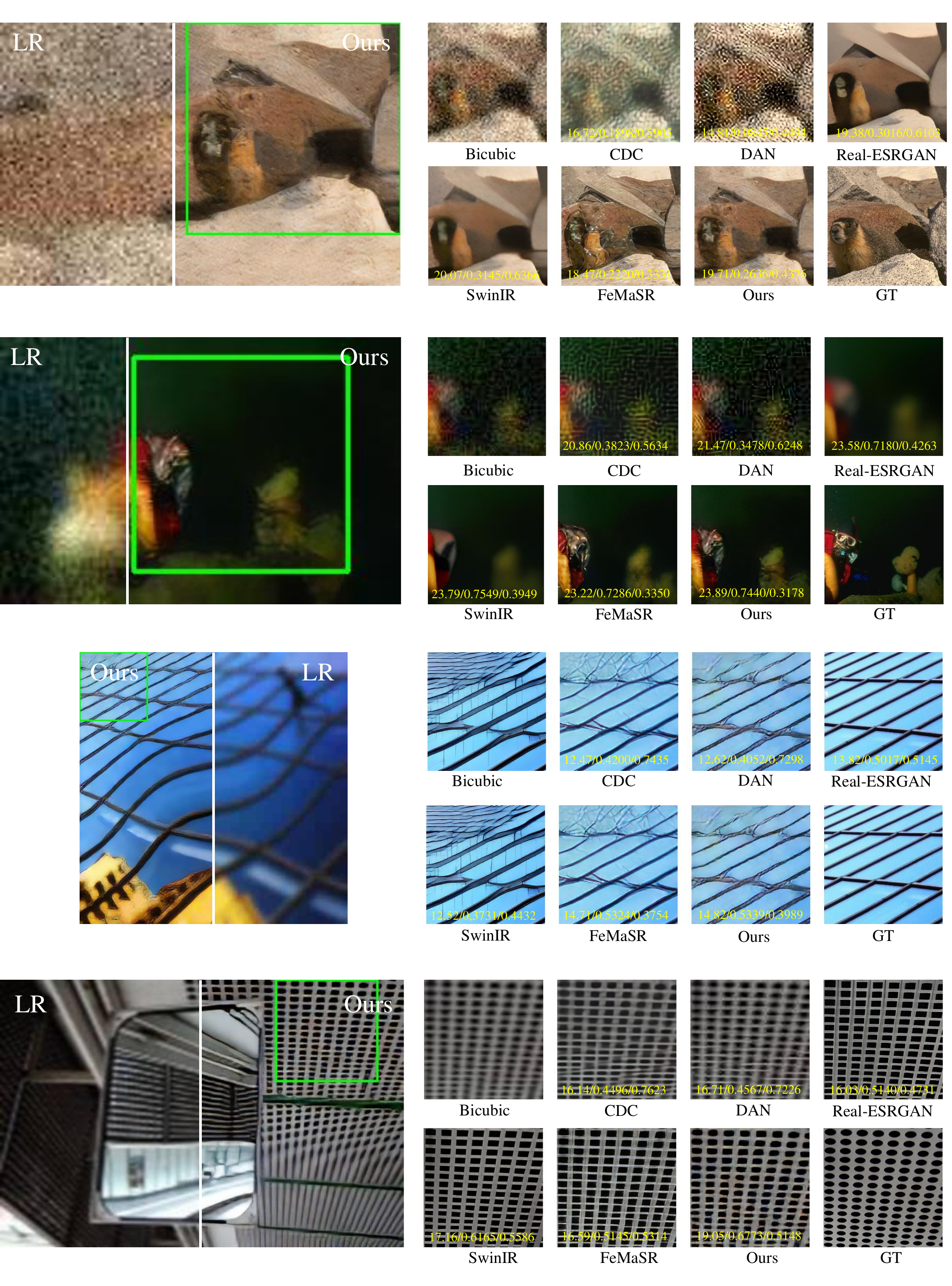}
    \vspace{-1cm}
	\caption{More results comparison in validation sets.}
	\label{fig:extr_sota_cmp1}
\end{figure*}

\begin{figure*}[htbp]
	\includegraphics[width=\linewidth]{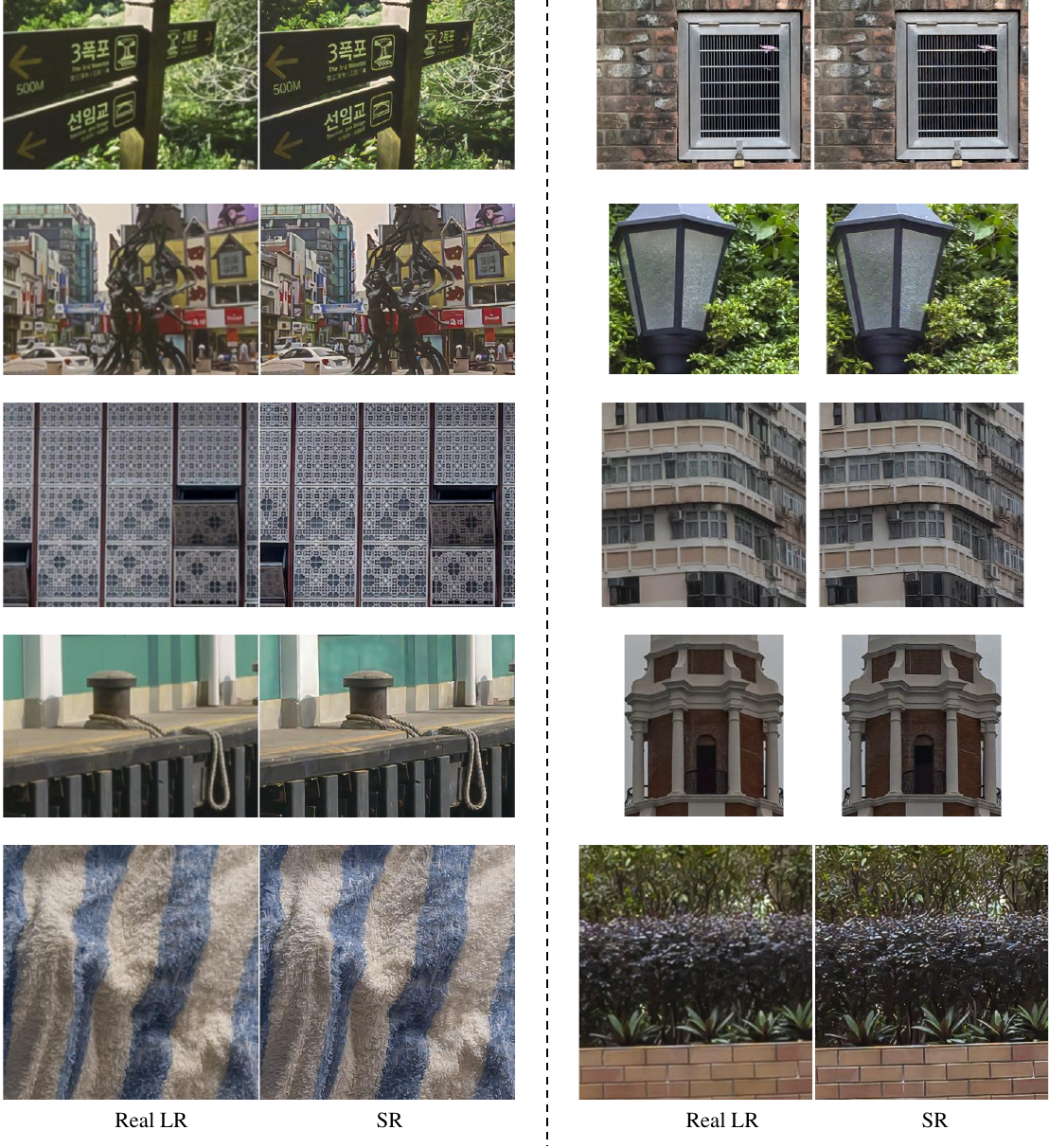}
	\caption{Super-resolution Results on real low-resolution images. (Zoom in for better performance)}
	\label{fig:real_sr}
\end{figure*}

\paragraph{\textbf{Extra Quanlitative Comparison}}
We show more qualitative results comparison in Fig.~\ref{fig:extr_sota_cmp1}.
\paragraph{\textbf{Results on Real LR Images}}
We also provide super-resolution results on real low-resolution images in Fig.~\ref{fig:real_sr}.
%\subsection{}

% \begin{figure*}[htbp]
%     \centering
% 	\includegraphics[width=\linewidth]{./extra_sota_cmp2.pdf}
% 		\caption{More results comparison in validation sets.}
% 	\label{fig:extr_sota_cmp2}
% \end{figure*}

% \subsection{Limitation and Discussion}
% First, 
% 	by observing the results, we find that RTCNet has some limitations when dealing with regular texture restoration, particularly in data types abundant in such textures, such as buildings(lines A.612-A.625). RTCNet tends to generate distorted edges often found in natural textures. This problem has also happened in the past codebook-based method, which we will explore in future work.
% Second, 
% 	we find that the improvement of RTCNet is more obvious in the more serious degradation than the lightly degraded samples(maybe no improvement in some light samples)(Fig.~\ref{fig:dtpm_femasr_cmp}). We surmise that the notable enhancements in heavily degraded data arise from the increased matching confusion, a scenario wherein RTCNet performs optimally. Conversely, lightly degraded samples with less confusion can also be handled well by previous methods causing marginal improvements. Even though both kinds of data are common in real-world applications, we contend that fixing complex degraded data possesses great challenges and value for super-resolution (SR) tasks.
% Besides, 
% 	in an intuitive sense, pre-training features tend to align more effectively with data with strong domain priors, suggesting that the application of the codebook method combined with pre-training strategies to specific types of data could be a direction worth investigation
% \subsection{}

\end{sloppypar}

\end{document}